\documentclass[journal]{IEEEtran}
\usepackage{xcolor}
\usepackage{cite}
\usepackage{hyperref}
\usepackage{amsmath}
\usepackage{amssymb}
\usepackage{pifont}
\usepackage{multirow}
\usepackage{array}
\usepackage{footnote}
\usepackage{booktabs}
\usepackage{tabularx}
\usepackage{array}
\usepackage{ragged2e}

\usepackage{moreverb,url}
\usepackage{algorithm,algorithmic}%
\usepackage[pdftex]{graphicx}
\ifCLASSINFOpdf
\else
\fi
\usepackage{amsfonts}

\usepackage{stfloats}
\begin{document}
%
\title{Large Language Models and Their Applications in Roadway Safety and Mobility Enhancement: A Comprehensive Review.}

%
%
%
\author{Muhammad Monjurul Karim , Yan Shi, Shucheng Zhang, Bingzhang Wang, Mehrdad Nasri, 
        Yinhai Wang\IEEEauthorrefmark{1}\thanks{All authors are with the Department of Civil and Environmental Engineering, University of Washington, Seattle, WA 98195, USA.}\thanks{\IEEEauthorrefmark{2}Email: mmkarim@uw.edu} \thanks{\IEEEauthorrefmark{1}Corresponding author. Email: yinhai@uw.edu}}

\maketitle


\begin{abstract}
Roadway safety and mobility remain critical challenges for modern transportation systems, demanding innovative analytical frameworks capable of addressing complex, dynamic, and heterogeneous environments. While traditional engineering methods have made progress, the complexity and dynamism of real-world traffic necessitate more advanced analytical frameworks. Large Language Models (LLMs), with their unprecedented capabilities in natural language understanding, knowledge integration, and reasoning, represent a promising paradigm shift. This paper comprehensively reviews the application and customization of LLMs for enhancing roadway safety and mobility. A key focus is how LLMs are adapted—via architectural, training, prompting, and multimodal strategies—to bridge the "modality gap" with transportation's unique spatio-temporal and physical data. The review systematically analyzes diverse LLM applications in mobility (e.g., traffic flow prediction, signal control, simulation) and safety (e.g., crash analysis, driver behavior assessment, rule formalization). Enabling technologies such as V2X integration, domain-specific foundation models, explainability frameworks, and edge computing are also examined. Despite significant potential, challenges persist regarding inherent LLM limitations (hallucinations, reasoning deficits), data governance (privacy, bias), deployment complexities (sim-to-real, latency), and rigorous safety assurance. Promising future research directions are highlighted, including advanced multimodal fusion, enhanced spatio-temporal reasoning, human-AI collaboration, continuous learning, and the development of efficient, verifiable systems. This review provides a structured roadmap of current capabilities, limitations, and opportunities, underscoring LLMs' transformative potential while emphasizing the need for responsible innovation to realize safer, more intelligent transportation systems.

\end{abstract}


\IEEEpeerreviewmaketitle

\section{Introduction}\label{}
Road transportation systems are integral to modern economies and societies, yet they persistently face significant challenges related to safety, efficiency, and sustainability \cite{10530969}. Traffic congestion results in substantial annual economic losses globally due to lost productivity, increased fuel consumption, and environmental impact. Moreover, roadway accidents remain a primary cause of preventable fatalities and injuries worldwide. While traditional transportation systems engineering, employing statistical models \cite{CHAND20215135}, physics-based simulations \cite{chongsim}, and rule-based control systems \cite{baskar2011rule}, has achieved notable progress, these methods often exhibit limitations in fully addressing the inherent complexity, dynamism, and uncertainty of real-world traffic environments. The heterogeneity of data sources, the demand for real-time decision-making, and the intricate interactions among vehicles, infrastructure, and human behavior necessitate more advanced and adaptive analytical frameworks.

The recent advancements in Artificial Intelligence (AI) \cite{jagatheesaperumal2024artificial}, particularly the significant progress in deep learning, have already initiated a transformation in transportation through enhanced capabilities in pattern recognition, prediction, and optimization. More recently, the emergence of Large Language Models (LLMs)—neural network architectures trained on extensive text corpora—signifies a potentially more profound paradigm shift. These models have demonstrated unprecedented capabilities in natural language understanding, knowledge integration, reasoning, and emergent abilities. Although initially developed for language-oriented tasks, LLMs are increasingly showing remarkable versatility for cross-domain applications, with transportation representing a particularly promising frontier.

The transportation domain offers unique opportunities for leveraging the inherent strengths of LLMs. Specifically, LLMs are adept at: (1) processing unstructured and multimodal data from diverse sources into a cohesive understanding; (2) reasoning about complex causal relationships and counterfactuals in traffic dynamics; (3) contextualizing decisions within broader socio-technical systems, including human behavior, regulations, and environmental factors; and (4) generating human-interpretable outputs that facilitate collaboration between AI systems and human stakeholders. Furthermore, the evolution towards Multimodal Large Language Models (MLLMs) \cite{liang2024survey}, capable of seamlessly processing information across text, images, video, and other modalities, further extends these capabilities within the sensor-rich transportation ecosystem.

The state-of-the-art in LLM development has advanced rapidly, with models such as GPT \cite{openai_gpt35turbo_2025}, Gemini \cite{team2023gemini}, Claude \cite{caruccio2024claude}, and Grok \cite{XAI2025} demonstrating increasingly sophisticated capabilities. Concurrently, the open-source community has produced progressively capable alternatives like DeepSeek \cite{liu2024deepseek}, Mistral \cite{jiang2023mistral7b}, Llama \cite{touvron2023llama}, Gemma \cite{team2024gemma}, Phi \cite{li2023textbooks}, and Qwen \cite{bai2023qwen}. This proliferation of both proprietary and open-source models has accelerated innovation and adaptation across various domains, creating a fertile ground for transportation-specific applications.

The potential for LLMs to revolutionize roadway safety and mobility is extensive. They can function as intelligent assistants for traffic engineers, automate complex analysis and design tasks, enable more natural human-machine interfaces, generate realistic scenarios for autonomous vehicle testing, improve traffic forecasting precision, optimize control strategies, enhance crash analysis from narrative reports, and facilitate knowledge transfer across transportation sub-domains. However, effectively harnessing these capabilities requires addressing significant challenges unique to the transportation context.

Despite their potential, directly applying general-purpose LLMs to sophisticated transportation problems encounters several barriers, often described as "modality gaps" between language-centric models and transportation's predominantly spatio-temporal, physical, and numerical data \cite{ling2023domain, ge2023openagi}. These gaps manifest as challenges in data grounding, spatial-temporal representation, computational efficiency for real-time requirements, verification and validation for safety-critical reliability, and explainability for transparent decision-making. Therefore, overcoming these obstacles necessitates thoughtful domain adaptation and specialized architectural innovations.

This review paper aims to address these needs by providing a comprehensive overview of the state-of-the-art in applying and customizing LLMs to enhance roadway safety and mobility. We synthesize findings from recent literature, detailing diverse methodologies and, critically, examining how LLMs are being adapted—through architectural modifications, specialized training techniques, and novel prompting strategies—to better suit transportation's unique data and challenges. Unlike many existing reviews that cover transportation broadly, our focus is deliberately on the interconnected yet distinct areas of safety and mobility. We recognize that LLM applications often differ significantly between these domains in terms of problem formulation, model customization, and computational demands, allowing for a more nuanced analysis and avoiding oversimplification. Furthermore, substantial attention is dedicated to cross-cutting enabling technologies and architectural approaches—such as edge computing, multimodal integration, domain adaptation techniques, and verification frameworks—that are pivotal for effective LLM deployment. By identifying domain-specific research gaps and challenges within both safety and mobility applications, we aim to provide transportation planners, engineers, and researchers with a clear roadmap of current capabilities, limitations, and future opportunities.

To tackle these aspects, the primary objectives of this paper are as follows:
\begin{itemize}
\item To systematically summarize how LLMs are customized and methodologies designed for transportation safety and mobility, aiming to develop a general framework for such adaptation, an aspect often neglected in existing surveys.
\item To systematically categorize and critically analyze diverse LLM application methodologies across the transportation safety-mobility spectrum.
\item To identify emerging trends, common architectural patterns, and novel uses of LLM capabilities (reasoning, multimodality, interaction).
\item To synthesize domain-specific research challenges, technical limitations, and potential risks in safety-critical deployments.
\item To uncover unexplored research opportunities and highlight promising directions for maximum impact on transportation outcomes.
\end{itemize}

By offering this synthesis, we aim to provide a clear, structured overview of the current landscape and future potential of LLMs in transforming transportation safety and mobility.

The remainder of this paper will further the discussion by presenting the following contents in sequence. Section 2 discusses related review works. Section 3 delves into the foundational concepts of LLMs relevant to transportation. Section 4 details LLM applications focused on enhancing mobility. Section 5 covers applications primarily aimed at improving roadway safety. Section 6 examines enabling technologies and cross-cutting frameworks. Section 7 discusses challenges and future research directions. Finally, Section 8 concludes the review.

\section{Related works}

The increasing prominence of Large Language Models (LLMs) has generated considerable interest in their application across various scientific and engineering domains, including transportation systems. In response to this trend, several review articles have recently assessed the role of LLMs within transportation contexts. While these existing surveys offer valuable perspectives, a critical examination reveals notable gaps, particularly concerning a focused and comprehensive synthesis of LLM applications specifically targeting the enhancement of roadway safety and mobility. This section, therefore, analyzes prominent existing surveys, highlighting their contributions and limitations to establish the necessity and unique contribution of the present work.

Several broad-scope surveys have sought to provide an extensive overview of LLM applications across the entirety of the transportation sector. For instance, Yan et al. \cite{yan2025large} present a systematic review covering autonomous driving, travel behavior, safety, aviation, and supply chains, synthesizing methods and identifying general challenges across this wide scope. Similarly, Nie et al. \cite{li2025applications} propose a unifying LLM4TR framework and taxonomy, reviewing methodological aspects across diverse applications from traffic prediction to urban mobility optimization. Wandelt et al. \cite{wandelt2024large} categorize over 130 studies within Intelligent Transportation Systems (ITS), encompassing areas from autonomous driving to tourism and traffic management. Furthermore, Mahmud et al. \cite{mahmud2025integrating} focus on integrating various LLMs with ITS, covering applications from traffic flow prediction to Vehicle to Everything (V2X) communication.

The principal limitation of these broad surveys \cite{yan2025large, li2025applications, wandelt2024large, mahmud2025integrating} stems from their extensive scope. While commendable for their breadth, this often results in a reduced depth of analysis within specific sub-domains. Consequently, roadway safety and mobility enhancement, though addressed, are frequently treated as components within a much larger ecosystem, lacking a dedicated, in-depth examination of the nuanced challenges, specific LLM adaptations, and unique opportunities pertinent to these interconnected areas. Moreover, some reviews \cite{yan2025large, li2025applications}, with literature searches concluding in late 2024, may not capture the most recent advancements in this rapidly evolving field. Others exhibit deficiencies in detailing deployment challenges \cite{li2025applications}, practical case studies \cite{wandelt2024large}, or relevant regulatory frameworks \cite{mahmud2025integrating}.

In contrast, other reviews adopt more focused methodologies, yet still present certain limitations. For example, Cui et al. \cite{cui2024survey} and Yang et al. \cite{yang2023llm4drive} provide valuable, in-depth surveys specifically on LLM applications within Autonomous Driving (AD), exploring aspects such as perception, planning, and human-vehicle interaction. Although AD is intrinsically linked to safety and mobility, the scope of these reviews is inherently constrained. They do not comprehensively cover non-AD aspects of roadway safety (e.g., traffic signal optimization, accident analysis, vulnerable road user safety) or broader mobility enhancement strategies beyond individual vehicles (e.g., network-level traffic management, public transport optimization). Cui et al. \cite{cui2024survey}, for instance, further omit discussion on integrating these technologies with broader ITS ecosystems.

Likewise, Zhang et al. (2024) \cite{zhang2024large} concentrate specifically on the application of LLMs for time-series forecasting tasks related to mobility, such as traffic flow, speed, and demand prediction. While forecasting constitutes a crucial component of mobility management, this review largely overlooks other vital aspects of mobility enhancement (e.g., dynamic route guidance generation, incident response strategies, traffic signal control optimization using LLM reasoning) and safety applications. Furthermore, it lacks comparative benchmarking and discussion on practical deployment challenges such as scalability and real-time processing.

These existing literature reviews tend to be either overly broad, thereby lacking the necessary depth concerning roadway safety and mobility, or excessively narrow, focusing on specific niches such as AD or forecasting without offering a holistic perspective on the interconnected safety and mobility domains. A significant research need exists for a review that specifically and comprehensively addresses the synergistic application of LLMs to both enhance roadway safety (encompassing accident analysis, prevention, infrastructure assessment, and post-incident response) and improve roadway mobility (including traffic flow optimization, congestion management, demand prediction, and multimodal integration) in an integrated manner. Crucially, prior reviews have not elaborately discussed how LLMs are being adapted for the transportation domain, nor have they provided a general framework for such customization or adequately covered the role of cross-cutting enabling technologies in conjunction with LLMs for this specific field. This paper aims to fill these identified gaps.

\section{Foundational Concepts: LLMs in the Transportation Context}
\label{sec:foundations}

To understand the application of LLMs in roadway safety and mobility, it is essential to grasp some fundamental concepts and adaptations relevant to this domain.

\subsection{Core LLM Concepts in the transportation context}
The advancement of Language models persisted with the emergence of the Transformer model in the year 2017 (\cite{vaswani2017attention}). LLMs are typically based on this Transformer architecture, which utilizes self-attention mechanisms to weigh the importance of different parts of the input sequence. This allows them to capture long-range dependencies effectively, crucial for understanding context in language and potentially in sequential transportation data. The transformer model has had a significant impact on the field of NLP and has played a crucial role in the development of language models such as Bidirectional Encoder Representations from Transformers (BERT) (\cite{devlin-etal-2019-bert}) and Generative Pre-trained Transformers (GPT) (\cite{floridi2020gpt, achiam2023gpt}).

\subsubsection{LLM training, fine-tuning, and domain adaptation:} The power of LLMs stems from pre-training on massive, diverse datasets (primarily text and code). This phase imbues the models with general knowledge, linguistic competence, and rudimentary reasoning abilities. 
Subsequently, fine-tuning adapts these pre-trained models to specific downstream tasks or domains \cite{zheng2023trafficsafetygpt, hanparameter}. This involves updating model parameters using domain-specific data. Techniques like Parameter-Efficient Fine-Tuning (PEFT), such as Low-Rank Adaptation (LoRA) \cite{hu2022lora}, allow adaptation with significantly fewer trainable parameters, making fine-tuning more computationally feasible. This has been successfully applied in fine-tuning LLMs for trajectory prediction \cite{lan2024traj, ren2024tpllm}, time-series forecasting \cite{guo2024towards, chang2023llm4ts}, analyzing driver distraction \cite{zhang2024integrating}, creating specialized models like TransGPT \cite{wang2024transgpt} and MetRoBERTa \cite{leong2024metroberta}, and adapting models for traffic signal control \cite{lai2023large}.

\subsubsection{Prompt Engineering}
Prompt engineering is the systematic process of designing and refining prompts—specific instructions or questions—to elicit desired outputs from LLMs. This involves crafting appropriate formats, phrasing, keywords, and contextual information to guide the model's generation process effectively. Given that many specialized transportation tasks lack the large, labeled datasets typically required for training LLMs from scratch or for extensive fine-tuning, prompt engineering has become an increasingly vital technique. It enables researchers and practitioners to leverage pre-trained LLMs for specific tasks such as information extraction, complex reasoning, and problem-solving within the transportation domain (\cite{dai2024large, li2024chatsumo, movahedi2024crossroads, kuftinova2024large, manas2024tr2mtl, guo2024towards, arteaga2025large, liang2024exploring, mao2023gpt, zhang2024trafficgpt, da2023llm, yildirim2024highwayllm, fang2025towards, sha2023languagempc, mumtarin2023large, sivakumar2024prompting, cai2025text2scenario, luo2024deciphering, hussien2025rag, syed2024benchmarking, masri2025large, jin2023time}).

Among various prompt engineering strategies, Chain-of-Thought (CoT) prompting has gained notable popularity. CoT techniques explicitly instruct the LLM to generate a series of intermediate reasoning steps before arriving at a final answer. This encourages the model to "think step-by-step," breaking down complex problems into more manageable parts, which can improve performance on tasks requiring logical deduction and provide transparency into the model's reasoning process. Such intermediate steps can often be inspected or verified, enhancing trust and interpretability. CoT is frequently employed to improve LLM performance in complex transportation tasks such as formalizing traffic rules \cite{manas2024tr2mtl}, generating explainable predictions \cite{guo2024towards}, or planning multi-step actions \cite{sha2023languagempc, lai2023large}.

The practical benefits of sophisticated prompt engineering, including CoT, are evident in recent transportation research. For instance, Zhen et al. \cite{zhen2024leveraging} explored the application of LLMs (GPT-3.5-turbo \cite{openai_gpt35turbo_2025}, LLaMA3-8B \cite{touvron2023llama}, and LLaMA3-70B \cite{touvron2023llama}) for classifying traffic crash severity from textual narratives generated from tabular crash data. The study demonstrated that employing CoT prompting significantly guided the models through a structured reasoning process about crash causality before inferring severity. Alongside CoT, meticulous prompt engineering (PE) was used to enhance task alignment, for example, by rephrasing sensitive labels like 'Fatal accident' to mitigate LLM alignment constraints. Their findings indicated that LLaMA3-70B\cite{touvron2023llama}) generally outperformed other models, especially in zero-shot scenarios, and that both CoT and specific PE techniques substantially improved classification performance and logical reasoning capabilities. Notably, the CoT approach provided valuable insights into the LLMs' reasoning, revealing their capacity to consider diverse factors such as environmental conditions, driver behavior, and vehicle characteristics when analyzing crash severity. This underscores the potential of advanced prompt engineering to unlock deeper analytical capabilities of LLMs in transportation safety.

\subsubsection{Retrieval-Augmented Generation (RAG):} Standard LLMs generate responses based solely on their internal, pre-trained knowledge, which can be outdated or lack domain-specific depth, potentially leading to inaccuracies or "hallucinations." Retrieval-Augmented Generation (RAG) addresses this limitation by integrating external knowledge retrieval into the generation process \cite{kenthapadi2024grounding, huang2025survey}. The RAG process typically involves three steps: (1) \textit{Retrieve:} Use the input prompt or query to search a relevant external knowledge base (e.g., a vector database containing traffic regulations, real-time incident reports, or technical manuals) and retrieve pertinent information snippets. (2) \textit{Augment:} Combine the original prompt with the retrieved contextual information. (3) \textit{Generate:} Feed this augmented prompt to the LLM, which then uses the provided context to generate a more accurate, relevant, and verifiable response \cite{hussien2025rag, liindustrial, you2025comprehensive, luo2025senserag, zhang2024ragtraffic}. This is particularly crucial for safety-critical transportation applications that demand responses grounded in up-to-date, factual, and domain-specific knowledge.

\subsubsection{LLM Agents and tool use} This powerful paradigm positions the LLM as a central coordinator or 'agent' that interacts with specialized external 'tools'. The LLM plans a sequence of actions to fulfill a user request, selects the appropriate tool (e.g., a traffic simulator, a database query function, a mathematical calculation module, a specific prediction model), formats the input for the tool, invokes it, and then interprets the tool's output to continue the plan or synthesize a final response. This allows leveraging the LLM's reasoning and flexibility while relying on the precision and domain-specificity of established tools. Examples include the Digital Traffic Engineer framework calling calculation and simulation tools \cite{panyam2025survey}, ChatSUMO interfacing with SUMO \cite{li2024chatsumo}, TrafficGPT \cite{zhang2024trafficgpt} and Open-TI \cite{da2024open}orchestrating various traffic analysis and simulation modules, and LA-Light using perception and decision tools for signal control \cite{wang2024llm}. Another integration involves code execution, where LLMs generate code as an intermediate product that is then executed to verify solutions, as demonstrated by Huang et al. (\cite{huang2024words}) in optimizing vehicle routing.

\subsection{Multimodal Large Language Models}
Multimodal Large Language Models (MLLMs) represent a significant advancement in deep learning, designed specifically to process, understand, and generate content across multiple data modalities—including text, images, audio, and video—rather than being confined solely to textual information \cite{wu2023multimodal}. The core architecture of MLLMs typically involves modality-specific encoders (e.g., a vision encoder for images or video frames, an audio encoder for sound, and a text encoder for language). The representations generated by these encoders are then integrated, often through a fusion module or by projection into a unified embedding space, enabling the model to perform cross-modal reasoning and generation \cite{wu2023multimodal}.

The transportation domain, characterized by its inherent multimodality involving visual data from cameras (traffic surveillance, dashcams, drones) and sensors (LiDAR, radar), alongside textual information (reports, signs), is particularly well-suited for the application of MLLMs. Vision-Language Models (VLMs), a prominent type of MLLM, are especially relevant, integrating computer vision capabilities with natural language processing \cite{cui2024survey}. These models typically combine a powerful vision encoder, such as CLIP \cite{radford2021learning} or Vision Transformers (ViT) \cite{dosovitskiy2020image}, with an LLM backbone \cite{wang2024transgpt, zhang2024integrating, pu2024autorepo}, allowing them to interpret visual scenes and articulate understanding through language.

Recent research demonstrates the growing application of MLLMs and VLMs to address complex challenges in roadway safety and mobility. For instance, Abu Tami et al. \cite{abu2024using} leverage MLLMs like Gemini-Pro-Vision \cite{team2023gemini} and Llava \cite{liu2023visual} to automatically analyze driving videos for safety-critical events. Their framework utilizes automated frame extraction and context-specific prompts within a multi-stage Question/Answering process to generate reliable insights about hazards, showcasing potential in zero-shot analysis of driving scenarios. Similarly, Zhang et al. \cite{zhang2025language} developed SeeUnsafe, an MLLM-based framework designed for interactive analysis of traffic accidents captured by surveillance cameras. By using task-specific multimodal prompts incorporating domain knowledge and visual cues (as depicted in Figure 1 of \cite{zhang2025language}), SeeUnsafe guides MLLMs to produce structured textual descriptions of accidents—including event classification, context, object details, and justification—and enables fine-grained visual grounding to identify involved road users. This approach aims to efficiently convert raw video footage into actionable safety intelligence.

Beyond direct event analysis from roadside or in-vehicle cameras, MLLMs are being integrated into broader safety management systems. Ahmed et al. \cite{ahmed2024detection} proposed a framework where computer vision models (YOLOv11s \cite{khanam2024yolov11}) first detect incidents like accidents or fires from drone/CCTV feeds. A VLM (Moondream2 \cite{korrapati_moondream2_2024}) then processes the visual data to generate detailed scene descriptions, which are subsequently refined by an LLM (GPT-4 Turbo \cite{openai_gpt35turbo_2025}) into concise summaries, actionable suggestions, and automated alerts for emergency services, thereby streamlining the entire incident response pipeline. Furthermore, MLLMs are being applied to understand nuanced driver behavior for safety improvement. Takato et al. \cite{takato2024multi} fine-tuned a Video-LLaMA \cite{zhang2023video} model on a custom dataset combining road-facing and driver-facing dashcam footage to enhance the analysis of driving events and generate detailed explanations suitable for driver coaching.
The exploration of VLMs in transportation extends to various other tasks, including general traffic scene understanding \cite{jain2024semantic, rivera2025scenario}, creating descriptive captions for accident videos \cite{xuan2024divide, dinh2024trafficvlm, qasemi2023traffic}, enabling visual question answering (VQA) systems tailored to driving contexts \cite{marcu2024lingoqa, rekanar2024optimizing}, and potentially supporting future cooperative driving applications \cite{you2024v2x}. Efforts are also underway to improve VLM performance in complex traffic environments by incorporating structured data representations like scene graphs \cite{lohner2024enhancing} or integrating outputs from instance segmentation models \cite{onsu2025leveraging}.

In essence, MLLMs provide powerful tools capable of bridging the gap between different data types prevalent in transportation. Their ability to jointly process and reason over visual and textual information opens up new avenues for enhancing situational awareness, automating analysis tasks, and ultimately contributing to safer and more efficient roadway systems.

\subsection{Specialized Architectures for Transportation Applications}
While standard Large Language Models possess remarkable capabilities honed on vast text corpora, they often encounter difficulties when directly applied to the unique data structures prevalent in transportation systems. Transportation data is typically numerical, multi-dimensional, and characterized by intricate spatial and temporal interdependencies that are fundamentally different from sequential language patterns. Recognizing this mismatch, researchers are actively developing specialized architectures and techniques to effectively bridge this gap and harness LLM strengths for transportation tasks. These specialized approaches can be broadly categorized as follows:

\subsubsection{Integration of Explicit Spatio-Temporal Modules}
A fundamental challenge in applying standard Large Language Models (LLMs) to transportation problems stems from their inherent design. Trained primarily on sequential text data, these models often lack the necessary inductive biases to effectively capture the complex, multi-dimensional, and physics-governed spatio-temporal dynamics inherent in transportation systems, such as traffic flow or vehicle interactions across networks \cite{rong2024large, rong2024edge, liu2024spatial, li2024urbangpt}. Transportation data is fundamentally different from language; it is often numerical and characterized by intricate interdependencies between locations (spatial) and points in time (temporal) that do not map neatly onto linear text sequences.

\begin{figure}[htbp]
    \centering
    \includegraphics[width=3.5in]{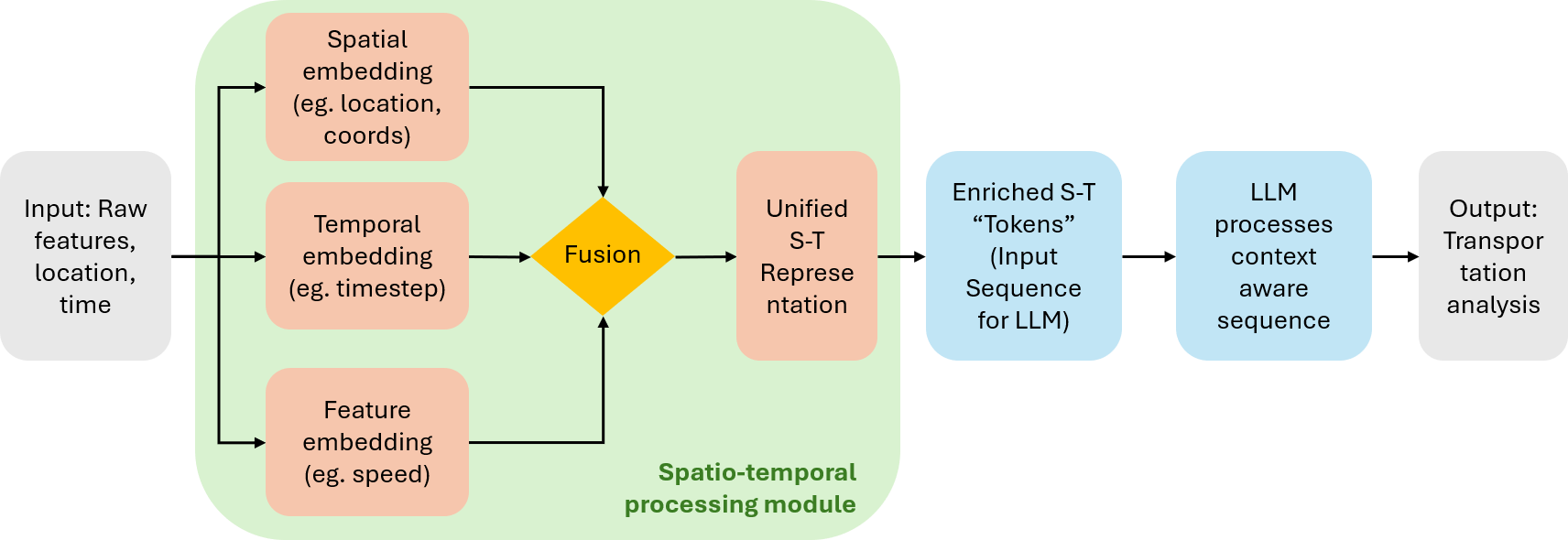} 
    \caption{Conceptual framework for augmenting Large Language Models (LLMs) with explicit Spatio-Temporal (S-T) modules for transportation applications.}
    \label{fig:st_module_framework}
\end{figure}

To bridge this gap, a prominent and effective strategy involves augmenting the standard LLM architecture by integrating dedicated modules specifically engineered to explicitly model these crucial spatio-temporal correlations \cite{rong2024large, rong2024edge, liu2024spatial, li2024urbangpt}. These specialized modules, often inspired by established techniques in spatio-temporal data analysis, can operate either upstream of the main LLM component (preprocessing and encoding the data) or work in tandem with it during the reasoning process. Figure \ref{fig:st_module_framework} provides a conceptual overview of this integration framework, illustrating how raw spatio-temporal data is processed by these specialized modules to create context-enriched inputs suitable for the LLM.

The figure illustrates the process of enhancing LLMs to handle complex transportation data. It begins with Raw Spatio-Temporal Data (comprising features, location, and time), which presents a challenge for standard LLMs due to their inherent lack of spatio-temporal inductive biases.

The proposed Augmentation Strategy involves integrating a dedicated Spatio-Temporal Processing Module. This module often takes the form of a spatial-temporal embedding layer, whose primary function is to transform raw spatio-temporal data points (e.g., traffic speed measured at sensor $X$ at time $T$) into dense vector representations, or embeddings. Within this module each incoming data point undergoes parallel encoding to generate these embeddings. The embeddings are designed to capture not only the specific features measured but also the critical context of where and when the measurement occurred, along with the underlying patterns and relationships between different locations and time steps.

\begin{itemize}
    \item Spatial Encoding typically involves embedding unique location identifiers, geographical coordinates, or incorporating graph structures representing the road network to capture location-based information. 
    \item Temporal Encoding often uses learnable embeddings or fixed sinusoidal functions to represent discrete time steps or cyclical patterns, thereby capturing time-based information.
    \item Feature Encoding processes the measured values themselves (e.g., speed, volume).
\end{itemize}

These spatial, temporal, and feature embeddings are then Fused, commonly through operations like addition or concatenation. This creates a Unified Spatio-Temporal (S-T) Representation for each data point, which effectively encapsulates the data's features along with its crucial spatio-temporal context.

This unified representation from the S-T module transforms each observation into Enriched S-T "Tokens." These tokens are analogous to word tokens in natural language processing but are imbued with rich spatio-temporal context, forming an input sequence suitable for the LLM.

The Large Language Model then processes this context-aware sequence of enriched tokens, enabling it to understand and reason about complex dependencies across the transportation network and over time.

Finally, the system produces the Output, which consists of transportation-specific analyses or predictions, such as traffic forecasts or safety assessments. This framework allows LLMs to be more effectively applied to roadway safety and mobility enhancement tasks.

Recent research provides several compelling examples of this integration strategy, representing specific implementations of the general framework shown in Figure \ref{fig:st_module_framework}. For instance, the LSGLLM-E \cite{rong2024large} and STGLLM-E \cite{rong2024edge} frameworks incorporate a distinct Spatio-Temporal Module (STM) featuring a novel and efficient spatial attention mechanism (STBMHSA) for spatial encoding. This mechanism is specifically optimized for resource-constrained edge devices, enabling the model to explicitly capture how traffic conditions propagate between nearby locations over time, a crucial capability for real-time, on-device transportation applications. Similarly, ST-LLM \cite{liu2024spatial} employs separate embedding layers dedicated to capturing spatial relationships and temporal periodicities, such as daily and weekly cycles. The outputs from these specialized layers are then fused before being processed by the main LLM. Taking a different approach, UrbanGPT \cite{li2024urbangpt} utilizes a multi-level Temporal Convolutional Network (TCN) to model temporal dynamics across various time scales. Complementing this, it employs a graph-independent method for spatial modeling, which offers enhanced flexibility compared to approaches reliant on predefined network graphs.

\subsubsection{Novel Input Representation and Tokenization Strategies} 

\begin{figure}[htbp]
    \centering
    \includegraphics[width=3.5in]{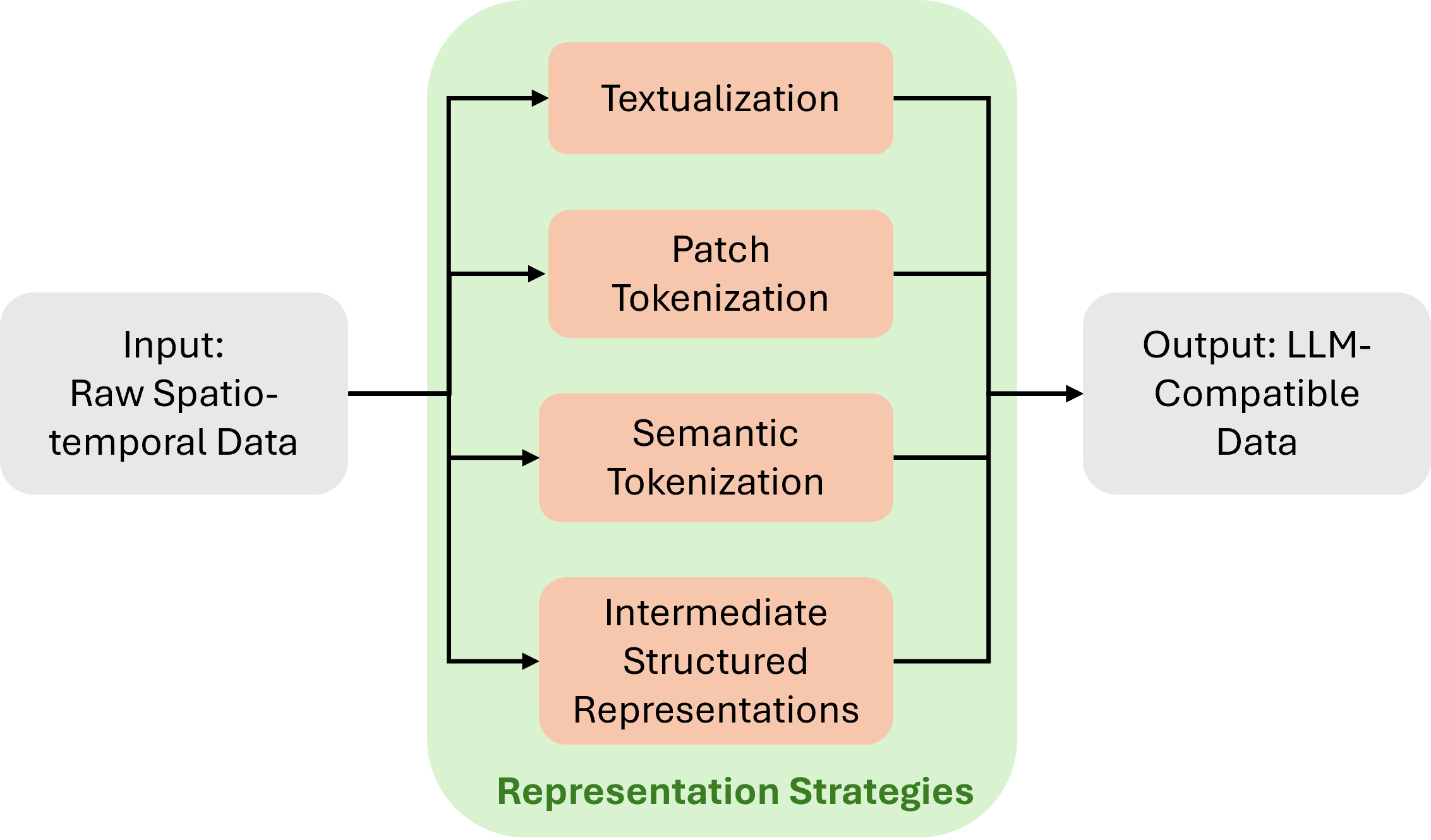} 
    \caption{Overview of input representation strategies for transforming spatio-temporal transportation data into LLM-compatible formats.}
    \label{fig:representation_framework}
\end{figure}

Standard Large Language Models (LLMs), predominantly based on the Transformer architecture \cite{vaswani2017attention} utilizing self-attention mechanisms, operate fundamentally on sequences of discrete tokens, $S = (t_1, t_2, ..., t_k)$. These tokens typically represent words or subwords derived from a predefined vocabulary, generated via algorithms like Byte Pair Encoding (BPE) or WordPiece. Each token $t_i$ is mapped to a high-dimensional vector embedding $e_i \in \mathbb{R}^d$, forming the input representation for the model. However, adapting numerical, continuous, and often high-dimensional spatio-temporal data, commonly found in transportation systems—which can be represented as a tensor $X \in \mathbb{R}^{N \times T \times F}$ (where $N$ is the number of spatial locations, $T$ is the number of time steps, and $F$ is the number of features per location/time)—into this discrete sequence format presents a significant challenge, often termed the ``modality gap.'' Bridging this gap requires innovative input representation techniques. Figure \ref{fig:representation_framework} illustrates the overall framework of input representation and tokenization strategies designed to transform structured spatio-temporal transportation data into LLM-compatible formats.

One major approach is 'textualization'. As employed by xTP-LLM \cite{guo2024towards}, this involves converting diverse inputs—such as historical traffic flow values, discrete spatial attributes (e.g., Point-of-Interest categories near sensors), continuous weather variables (e.g., temperature, precipitation), and temporal markers (date, time, holiday indicators)—into structured natural language prompts. For instance, a data point might be translated to: ``Sensor $i$, located near a `shopping mall', recorded a flow of 50 vehicles/hour at 16:00 on 2024-10-26 (weekday), with temperature 15\,$^{\circ}\text{C}$ and `rainy' conditions.'' This strategy leverages the LLM's inherent natural language processing (NLP) capabilities and can enhance the explainability of predictions. However, textualization inherently involves discretization or binning of continuous values, potentially leading to information loss (quantization error), and can generate excessively long input sequences, increasing computational demands.

Other strategies focus on novel 'tokenization' methods tailored for time-series data. STGLLM-E \cite{rong2024edge} adapts the concept of ``patching'' from Vision Transformers (ViTs) \cite{dosovitskiy2020image}. The time-series data for a location, $X_i \in \mathbb{R}^{T \times F}$, is segmented into non-overlapping (or overlapping) patches $P_{i,j} \in \mathbb{R}^{L \times F}$, where $L$ is the patch length. Each patch is then typically flattened and linearly projected into the LLM's embedding dimension $d$, effectively treating each patch as a ``token'': $e_{i,j} = \text{Linear}(P_{i,j}) \in \mathbb{R}^d$. In contrast, ST-LLM \cite{liu2024spatial} adopts a coarser granularity, uniquely treating the \textit{entire} time-series matrix for a single location over a past window, $X_i^{t-T_{past}:t} \in \mathbb{R}^{T_{past} \times F}$, as a single conceptual ``token.'' This requires a specialized embedding function $E: \mathbb{R}^{T_{past} \times F} \rightarrow \mathbb{R}^d$ to map this high-dimensional object into the LLM's input space. TIME-LLM \cite{jin2023time} pursues a semantic tokenization approach, more akin to Symbolic Aggregate Approximation (SAX) or other time-series discretization techniques. It reprograms raw time-series values or subsequences into descriptive textual tokens (e.g., `short up', `steady down', `sharp drop') based on predefined rules or potentially learned patterns (e.g., via vector quantization), abstracting the numerical details into qualitative descriptors.

A distinct paradigm involves using 'intermediate structured representations'. Here, the LLM does not directly process the raw spatio-temporal data $X$. Instead, it interprets a natural language \textit{request} or context description and translates it into a structured format (e.g., JSON, domain-specific codes). LCTGen \cite{tan2023language}, for example, uses an LLM to generate interaction codes, vehicle codes, and map codes, which subsequently guide downstream, domain-specific models (like traffic simulators or predictive algorithms) that \textit{do} operate directly on the numerical data. In this setup, the LLM functions more as a high-level planner or translator.

The overarching goal of these diverse methods is to effectively translate the rich, continuous, and structured information inherent in transportation data ($X \in \mathbb{R}^{N \times T \times F}$) into a sequential, often discrete, format ($S$ or a sequence of specialized embeddings $e_i$) that LLMs can process, thus bridging the fundamental modality gap between complex spatio-temporal dynamics and sequence-based language modeling. The choice among these techniques involves trade-offs between information fidelity, computational complexity, model adaptability, and the desired level of interpretability.

\subsubsection{Partially Frozen Model Adaptation} 

Fully fine-tuning a massive pre-trained LLM, involving updating all its parameters $\theta$ based on the transportation task's loss function $L_{task}$ (i.e., $\theta_{new} = \theta_{pre} - \eta \nabla_{\theta} L_{task}$, where $\eta$ is the learning rate), can be computationally prohibitive for relatively small or domain-specific transportation datasets. Furthermore, it risks ``catastrophic forgetting,'' where the gradient updates optimized for $L_{task}$ overwrite the parameters crucial for the general knowledge acquired during pre-training, diminishing the model's foundational capabilities. Conversely, using a completely frozen LLM (where only a minimal task-specific head $\phi$ might be trained, keeping the vast pre-trained parameters $\theta_{pre}$ fixed) often fails to capture the specific nuances and complex dependencies inherent in transportation data, limiting adaptive performance.

`Partially frozen models' offer a pragmatic compromise by partitioning the model's parameters $\theta$ into a frozen set $\theta_{frozen}$ and a tunable set $\theta_{tunable}$ ($\theta = \theta_{frozen} \cup \theta_{tunable}$). During adaptation, only the tunable parameters are updated: $\theta_{tunable, new} = \theta_{tunable, pre} - \eta \nabla_{\theta_{tunable}} L_{task}$. This approach often follows the hypothesis that lower layers of deep networks learn more general features, while upper layers specialize in higher-level, potentially task-specific patterns. Techniques like Partially Frozen Attention (PFA), as utilized in ST-LLM \cite{liu2024spatial}, exemplify this by keeping the parameters of the initial Transformer layers (potentially including both the self-attention sub-layers with weights for Query ($W^Q$), Key ($W^K$), Value ($W^V$), and Output projection ($W^O$), and the feed-forward network sub-layers with weights $W_1, W_2$) frozen. This preserves the foundational knowledge. Subsequently, specific components in the upper layers, such as the multi-head attention mechanisms responsible for capturing complex spatio-temporal interactions, are designated as $\theta_{tunable}$ and fine-tuned on the transportation task. This targeted adaptation allows the model to specialize its pattern recognition capabilities for spatio-temporal dependencies without the full computational cost or forgetting risk of complete fine-tuning, striking an effective balance between leveraging pre-trained power and achieving domain-specific performance.

Parameter-Efficient Fine-Tuning (PEFT) techniques represent a closely related and increasingly popular philosophy. Methods like Low-Rank Adaptation (LoRA) \cite{hu2022lora}, utilized in several transportation LLM studies \cite{lan2024traj, guo2024towards, ren2024tpllm}, also adapt only a small fraction of parameters. Specifically, LoRA \cite{hu2022lora} keeps the original pre-trained weight matrices $W_0 \in \mathbb{R}^{m \times n}$ frozen and injects trainable low-rank matrices $A \in \mathbb{R}^{m \times r}$ and $B \in \mathbb{R}^{r \times n}$ (where the rank $r \ll \min(m, n)$) into specific layers (often the attention mechanism weights). The effective weight update is represented by the low-rank product $AB$, so the adapted weight becomes $W_{adapted} = W_0 + AB$. Only parameters in $A$ and $B$ are trained, drastically reducing the number of tunable parameters from $m \times n$ to $(m+n)r$, thereby significantly lowering memory requirements and computational cost during fine-tuning while often maintaining performance comparable to full fine-tuning. These PEFT methods align with the partial freezing strategy by selectively modifying parts of the model to balance adaptation and knowledge preservation.

\subsubsection{LLM-as-Orchestrator Integration Frameworks}
\label{sec:llm_orchestrator} 

A prominent approach to address LLMs' limitations in precise numerical computation, simulation execution, and specialized algorithm application is the development of an architectural paradigm that positions the LLM as an orchestrator or intelligent interface. This approach, aligned with the broader trend of tool-augmented LLMs and agent-based systems, leverages the LLM as a central reasoning component that interacts with external capabilities. Instead of performing the core transportation modeling or analysis task itself, the LLM manages a suite of external, specialized tools, models, or Application Programming Interfaces (APIs). These external components can range from established transportation planning software and simulation engines to pre-trained domain-specific models (sometimes referred to as Traffic Foundation Models or TFMs).

The typical interaction involves the LLM receiving a user request, often in natural language (e.g., ``Find the optimal signal timing for the intersection of 5th Ave and Pine St during evening peak hours using Webster's method, considering current demand patterns''). The LLM then orchestrates the process by interpreting the request, selecting and invoking the necessary external components, and synthesizing their outputs into a final response. This general workflow, including key notations, inputs, and outputs, is formally described by Algorithm \ref{alg:llm_orchestrator}.

\begin{algorithm}[htbp] %
    \caption{LLM-as-Orchestrator Framework}
    \begin{algorithmic} 
        \STATE \textbf{Notation}: 
        \STATE $R_{user}$: The user's request in natural language or structured format
        \STATE $\mathcal{T}$: Library of available external tools, models, or APIs
        \STATE $S$: Set of subtasks decomposed from $R_{user}$, indexed by $s_i$
        \STATE $t_i$: An external tool selected from $\mathcal{T}$ to execute subtask $s_i$
        \STATE $p_i$: Parameters formatted for tool $t_i$ to execute subtask $s_i$
        \STATE $r_i$: Intermediate result obtained from executing tool $t_i$ with parameters $p_i$
        \STATE $R_{intermediate}$: Collection of all intermediate results $\{r_i\}$
        \STATE $A_{final}$: The final synthesized response to the user
        \vspace{0.03in}
    
        \STATE \textbf{Input}: User request $R_{user}$, Tool library $\mathcal{T}$
        \STATE \textbf{Output}: Final response $A_{final}$
        \vspace{0.03in}
    
        \STATE $S \leftarrow \text{LLM.Decompose}(R_{user})$   
        \STATE $R_{intermediate} \leftarrow \emptyset$
        
        \FOR{each sub-task $s_i \in S$}
            \STATE $t_i \leftarrow \text{LLM.SelectTool}(s_i, \mathcal{T})$
            \STATE $p_i \leftarrow \text{LLM.FormatParams}(s_i, t_i, R_{user})$
            \STATE $r_i \leftarrow \text{ExecuteTool}(t_i, p_i)$ 
            \STATE $R_{intermediate} \leftarrow R_{intermediate} \cup \{r_i\}$
        \ENDFOR
        
        \STATE {$A_{final} \leftarrow \text{LLM.Synthesize}(R_{intermediate}, R_{user})$}
    \end{algorithmic}
    \label{alg:llm_orchestrator} 
\end{algorithm}

The workflow, as outlined in Algorithm \ref{alg:llm_orchestrator}, commences with the LLM utilizing its `LLM.Decompose` capability to break down the initial user request ($R_{user}$) into a structured set of subtasks ($S$). Following this decomposition, the algorithm iterates through each subtask $s_i$. Within this loop, the LLM employs `LLM.SelectTool` to identify the most suitable external tool $t_i$ from the provided library $\mathcal{T}$. Subsequently, `LLM.FormatParams` prepares the necessary parameters $p_i$ for the chosen tool, drawing information from the subtask $s_i$ and the original request $R_{user}$. The `ExecuteTool` function is then invoked with $t_i$ and $p_i$, producing an intermediate result $r_i$. These individual results are aggregated into a collection $R_{intermediate}$. The process culminates with the `LLM.Synthesize` function, which processes $R_{intermediate}$ (and often the original $R_{user}$ for context) to generate the comprehensive final response $A_{final}$.

Several recent systems exemplify this framework. \textbf{TrafficGPT} \cite{zhang2024trafficgpt} demonstrates an LLM orchestrating TFMs for tasks like data querying, visualization generation, SUMO simulation control, and invoking classical algorithms (e.g., Webster's method), closely following the conceptual steps detailed in Algorithm \ref{alg:llm_orchestrator}. \textbf{VistaGPT} \cite{tian2023vistagpt} employs an LLM within its AutoAuto component, functioning as a meta-controller that automatically selects and integrates modular Transformers (for perception, prediction, planning) from a library to compose end-to-end autonomous driving pipelines. Similarly, \textbf{Open-TI} \cite{da2024open} provides an augmented LLM framework designed for complex traffic analysis; it maintains a dialogue with the user and intelligently calls upon external tools like osm2gmns \cite{lu2023virtual} (map processing), grid2demand \cite{luo2024grid2demand} (demand generation), DLSim \cite{DLSimMRM2023} (microscopic simulation), and Libsignal \cite{wei2023libsignal} (signal control) to fulfill tasks specified via natural language.

These orchestration frameworks strategically leverage the LLM's strengths in understanding ambiguous human language, reasoning about task structure, and planning execution flows. Simultaneously, they rely on validated, domain-specific tools for the computational heavy lifting, adherence to physical laws (in simulation), mathematical precision, and specialized knowledge representation. This separation of concerns aims to create powerful, flexible transportation analysis systems that are potentially more trustworthy, as the critical computations are handled by dedicated, often verifiable, external modules rather than solely by the LLM's internal, sometimes less transparent, mechanisms. This orchestration paradigm, therefore, represents a critical approach for building robust, reliable, and versatile LLM-powered systems capable of addressing complex transportation challenges by effectively combining LLM intelligence with specialized computational tools.

\section{LLM Applications for Mobility Enhancement}
LLMs offer significant potential to improve the efficiency, reliability, and user experience of transportation systems. This section reviews key applications focused on mobility enhancement.

\subsection{Traffic Flow Prediction and Forecasting}

Accurate traffic flow prediction and forecasting are fundamental for effective traffic management, route guidance, and congestion mitigation, directly contributing to enhanced urban mobility. These capabilities directly contribute to enhanced urban mobility and more efficient transportation networks. Formally, traffic flow prediction is defined as the task of forecasting future traffic states (e.g., flow, speed, or occupancy) at specific locations and times, based on historical observations and potentially external contextual data.

Mathematically, the general forecasting problem can be expressed as learning a mapping function $\pmb{f}$ such that:
\begin{equation}
    \hat{\pmb{X}}_{t+1:t+H} = f(\pmb{X}_{t-T+1:t}, \pmb{E}_{t-T+1:t})
\end{equation}
where $\pmb{X}_{t-T+1:t}\in\mathbb{R}^{T \times N})$ denotes the historical traffic data from $N$ locations over $T$ time steps, $\pmb{E}_{t-T+1:t}$ represents external factors (such as weather or events), and $\hat{\pmb{X}}_{t+1:t+H}$ is the predicted traffic clow for the next $H$ time steps.
    
Traditional approaches, encompassing statistical models (e.g., ARIMA, SARIMA), classical machine learning (e.g., SVR \cite{drucker1997support}), and early deep learning methods (e.g., Recurrent Neural Networks like LSTM \cite{hochreiter1997long} and GRU \cite{cho2014learning}), often struggle with capturing the complex, non-linear spatio-temporal dependencies inherent in traffic systems. These dependencies manifest as correlations between traffic conditions at nearby locations (spatial) and across consecutive time intervals (temporal), often influenced by network topology and time-varying factors. Additionally, integrating diverse, often unstructured, external factors like public events, accidents, or weather conditions poses a significant challenge for purely numerical models. Large Language Models (LLMs) are emerging as powerful tools to address these limitations, leveraging their sophisticated sequence modeling capabilities and capacity for semantic understanding. Research in this area explores various strategies, ranging from adapting LLM architectures specifically for spatio-temporal data patterns to leveraging their semantic capabilities for contextual enrichment and employing large pre-trained models with minimal fine-tuning through innovative prompting or input reprogramming techniques.

One significant line of research involves adapting established LLM architectures, initially designed for natural language, to the nuances of traffic data. Jin et al. \cite{JIN2021115738} pioneered this direction by adapting the BERT (Bidirectional Encoder Representations from Transformers) \cite{devlin-etal-2019-bert} architecture into TrafficBERT for long-range traffic flow forecasting. Pre-trained using a Masked Language Model-style objective on large-scale, aggregated traffic speed datasets from multiple locations, TrafficBERT utilizes the transformer's core multi-head self-attention mechanism. This mechanism allows the model to weigh the importance of different time steps when representing a specific point in the sequence. TrafficBERT incorporates specific traffic value embeddings alongside positional embeddings and weekday embeddings, enabling it to capture complex temporal patterns. This approach demonstrated superior performance compared to traditional time-series models (ARIMA) and earlier deep learning methods (SAE, LSTM \cite{hochreiter1997long}, GRU\cite{cho2014learning}). Crucially, it highlighted the benefit of transfer learning – leveraging knowledge gained from pre-training on diverse traffic datasets – for enhanced robustness and generalizability when fine-tuning for specific road networks or conditions.

More recent work has focused on adapting generative pre-trained models like those in the GPT (Generative Pre-trained Transformer) family \cite{floridi2020gpt, achiam2023gpt}, which are typically pre-trained using auto-regressive objectives (predicting the next token). Liu et al. \cite{liu2024spatial} introduced the Spatial-Temporal Large Language Model (ST-LLM), modifying GPT-2. Uniquely, it treats the entire historical time-series data for a single location (a vector or matrix) as a distinct input ``token,'' requiring specialized embedding techniques. It incorporates domain-specific embeddings reflecting day-of-week, hour-of-day, and spatial proximity (potentially derived from graph distances or adjacency matrices). Their partially frozen attention strategy (discussed in the previous section) preserves foundational knowledge from the language pre-training while adapting specific upper attention layers to capture traffic dynamics effectively. Similarly, Ren et al. \cite{ren2024tpllm} proposed TPLLM, which first uses Convolutional Neural Networks (CNNs) – effective for extracting local patterns in sequences – to generate sequence embeddings, and Graph Convolutional Networks (GCNs) \cite{zhang2019graph} – designed to operate on graph-structured data – for spatial graph structure embedding. These rich representations are then fed into a GPT-2 backbone that is efficiently fine-tuned using Low-Rank Adaptation (LoRA) \cite{hu2022lora}. Both ST-LLM and TPLLM showed strong performance, particularly excelling in few-shot learning scenarios (i.e., adapting to new locations or prediction tasks with very limited training examples), indicating LLMs' capacity to generalize effectively from sparse traffic data. Further refining architectural integration, Li et al. \cite{li2024urbangpt} developed UrbanGPT. This model combines a spatio-temporal dependency encoder, notably using temporal convolutions (TCNs) \cite{lea2016temporal} which capture temporal dependencies effectively without recurrence and without requiring pre-defined graph structures (enhancing flexibility), with an instruction-tuning paradigm. Instruction-tuning involves fine-tuning the LLM to follow natural language commands describing the desired task. By explicitly aligning textual instructions with learned spatio-temporal representations, UrbanGPT achieved excellent zero-shot (i.e., performing tasks without any specific examples) and cross-city prediction performance on diverse urban datasets.

Beyond direct architectural adaptation for the core forecasting task, LLMs' natural language processing strengths are being harnessed to incorporate rich contextual information often missing from purely numerical models. Huang et al. \cite{huang2024enhancing} proposed a novel hybrid method where the LLM is \textit{not} used for direct forecasting. Instead, it specifically processes textual descriptions of relevant contextual factors – regional events (e.g., weather forecasts, public holidays) and local incidents (e.g., concerts, sporting events, accidents). The LLM generates dense vector representations, or semantic embeddings, capturing the meaning of this text. These embeddings are then integrated, typically via concatenation or attention mechanisms, as auxiliary features within conventional spatio-temporal forecasting models (e.g., GCN-based predictors): $Input_{ST\_Model} = [Features_{numeric}, Embed_{LLM}(Text_{context})]$. This approach significantly improved prediction accuracy in a New York City bike-sharing case study, particularly during special circumstances where context is critical. It demonstrates an effective way to fuse qualitative textual context with quantitative historical data without incurring the high computational cost or potential numerical instability of using an LLM for the entire end-to-end forecasting process.

Another promising direction involves leveraging the vast knowledge embedded in large, pre-trained LLMs with minimal or no task-specific fine-tuning, capitalizing on their emergent capabilities. Jin et al. \cite{jin2023time} introduced TIME-LLM, which ingeniously reformulates the forecasting problem. It first "reprograms" numerical time series data into a sequence of discrete textual tokens representing learned patterns or "prototypes" (e.g., symbolic representations like ``short up'' or ``steady decline''). Using a Prompt-as-Prefix (PaP) technique – prepending carefully crafted text prompts to the input sequence – they guide a frozen, general-purpose LLM (like Llama-7B \cite{touvron2023llama}) to perform forecasting by predicting the next text prototype, which is then decoded back to a numerical value. This achieved strong few-shot and zero-shot results without modifying the core LLM parameters. Chang et al. \cite{chang2023llm4ts} developed LLM4TS using a two-stage fine-tuning approach (first aligning time series representations with the LLM's embedding space, then fine-tuning for the forecasting task) combined with novel temporal encoding strategies, also demonstrating excellent few-shot performance. Furthermore, the concept of foundation models specifically for time series has emerged, exemplified by Lag-Llama \cite{rasul2023lag}. This model is pre-trained on a massive corpus of diverse time series data (not just language or traffic). It can perform zero-shot forecasting across various domains, including traffic, by taking lagged values of the target series (e.g., $y_{t-L}, ..., y_{t-1}$, where $L$ is the context length) as input. A key advantage is its ability to provide probabilistic forecasts, outputting parameters of a distribution (e.g., mean and variance, or quantiles) rather than just a single point estimate, thereby offering valuable insights into prediction uncertainty.

\subsection{Traffic Data Analysis and Decision Support}

The sheer volume and complexity of transportation data, encompassing real-time sensor readings, historical trends, infrastructure details, incident reports, and user feedback, present significant challenges for analysis and effective decision-making. Accessing and interpreting this data often requires specialized technical skills, creating barriers for many transportation professionals and stakeholders. Large Language Models (LLMs) offer a paradigm shift in interacting with and extracting insights from this data, leveraging their natural language processing capabilities and reasoning abilities to provide intuitive interfaces and enhance analytical workflows.

One significant application area is the development of natural language interfaces for querying transportation databases. Traditional database interaction requires structured query languages (e.g., SQL), limiting accessibility. LLMs are being explored to bridge this gap by translating natural language questions into executable database queries. This is formally a Text-to-SQL mapping problem where LLM models the function $f: Q_{NL} \rightarrow Q_{SQL}$, mapping natural language queries ($Q_{NL}$) to executable SQL statements ($Q_{SQL}$).  While challenges remain, particularly with large-scale, noisy databases and complex queries requiring external knowledge, as highlighted by benchmarks like BIRD \cite{li2023can}, progress is being made through specialized pre-training frameworks like STAR \cite{cai2022star} aimed at improving context-dependent Text-to-SQL parsing.

In the transportation domain, several innovative systems exemplify this approach. Padoan et al. \cite{padoan2024mobility} presented Mobility ChatBot, an LLM agent architecture (using Langchain \cite{langchain}, Llamaindex \cite{llamaindex}, GPT-4-Turbo \cite{openai_gpt35turbo_2025}) specifically designed to allow users without SQL expertise to query structured mobility datasets using natural language. The agent interprets user intent, generates appropriate SQL queries, executes them, performs basic calculations or generates plots, and presents results conversationally. Similarly, Wang et al. \cite{karim2024} developed TP-GPT, an intelligent online chatbot for customized transportation surveillance and management, powered by a large real-time traffic database. Their framework leverages LLMs to generate accurate SQL queries and natural language interpretations, employing techniques like transportation-specialized prompts, Chain-of-Thought reasoning, few-shot learning, multi-agent collaboration, and chat memory. Further pushing the boundaries of automation and accessibility, Yang et al. \cite{yang2025independent} introduced Independent Mobility GPT (IDM-GPT), a novel multi-agent LLM framework. IDM-GPT uses five specialized agents to manage query validation, prompt optimization, secure database interaction, ML model selection/execution, and result evaluation, enabling users without transportation or ML backgrounds to obtain analysis and tailored suggestions via natural language. A key benefit highlighted by IDM-GPT is enhanced data privacy by minimizing direct human access to sensitive information. Collectively, these systems democratize data access, supporting more informed decision-making for planners, city officials, and other stakeholders.

Beyond direct data querying, LLMs contribute to structuring complex domain knowledge for better decision support. Tupayachi et al. \cite{tupayachi2024towards} explored using LLMs to automate the construction of scientific ontologies from technical documents and research articles. In their case study on optimizing intermodal freight transportation using datasets like Freight Analysis Framework (FAF) \cite{FAF5_2024} and Freight and Fuel Transportation Optimization Tool (FTOT) \cite{FTOT2022}, the AI-generated ontologies (in OWL format) facilitated data modeling, improved data integration for simulations (freight, traffic, emissions), and guided the design of decision support strategies. This approach significantly accelerates the traditionally time-consuming and expertise-intensive process of ontology engineering, paving the way for more sophisticated data-driven decision support systems in complex urban environments.

LLMs can also augment specific data processing and analysis tasks within larger workflows. Handling missing data, for instance, is a persistent problem. Zhang et al. \cite{zhang2024semantic} proposed the Graph Transformer-based Traffic Data Imputation (GT-TDI) model, which leverages GNN and Transformer architectures for imputation based on spatio-temporal semantics. Critically, they integrated an LLM and prompt engineering to provide a natural language interface to this sophisticated system. This allows users to interact with and request imputation results using plain language, enhancing the usability and interpretability of advanced data imputation techniques without requiring deep technical understanding.

Perhaps the most transformative potential lies in frameworks where LLMs act as central orchestrators for complex analytical workflows, integrating multiple data sources and specialized tools. Zhang et al. \cite{zhang2024trafficgpt} introduced TrafficGPT, which employs an LLM to manage interactions with various specialized traffic foundation models (TFMs). Given a high-level request, TrafficGPT decomposes the task, selects and invokes appropriate TFMs (for data extraction, processing, visualization, simulation control via SUMO, or signal optimization), manages dialogue, and synthesizes results. It demonstrated capabilities in large-scale OD flow analysis, network performance evaluation, and automated signal control optimization. The aforementioned IDM-GPT \cite{yang2025independent} also fits this paradigm, using its multi-agent system to manage a workflow encompassing querying, ML model execution, and evaluation. These orchestration frameworks showcase how LLMs can coordinate domain-specific tools to provide comprehensive, accessible, and powerful platforms for traffic analysis, management, and decision support.

These principles find specific and impactful applications within public transit systems. Wang and Shalaby \cite{wang2024leveraging} demonstrated how LLMs can serve as intermediaries between natural language content and transit databases. They presented practical tools like Tweet Writer (automating social media alerts), Trip Advisor (providing personalized, accessibility-aware journey recommendations), and Policy Navigator (offering clear answers to policy queries via RAG). These leverage LLM reasoning and retrieval techniques to improve customer experience and staff efficiency. Jonnala et al. \cite{jonnala2025exploring}, analyzing GTFS data \cite{kingcounty_gtfs_2025}, found LLMs possess significant latent transit knowledge and perform well on simpler retrieval tasks, though complex multi-file queries remain challenging. These studies underscore the potential for LLMs to revolutionize public transit planning, communication, and operations, despite current limitations requiring careful engineering.

\subsection{Traffic Signal Control and Optimization}

Optimizing traffic signal control (TSC) is crucial for enhancing urban mobility by reducing congestion, vehicle delays, and stops, thereby improving overall traffic flow, particularly at intersections and coordinated arterials. Large Language Models (LLMs) are emerging in this domain, moving beyond traditional and purely Reinforcement Learning (RL) approaches. Leveraging their advanced reasoning and generalization, LLMs are being explored to assist traffic engineers, act as decision-making agents, and enhance the robustness and adaptability of existing control systems.

Initially, LLMs were conceptualized as intelligent assistants to human traffic engineers. Dai et al. \cite{dai2024large} introduced the concept of Digital Traffic Engineers (DTEs), utilizing GPT-4 to interpret natural language instructions. These DTEs demonstrated proficiency automating complex tasks like analyzing network topology from simulation data (SUMO), calculating signal timings using established methods (e.g., Webster's formula), and even facilitating the design of advanced RL-based adaptive controllers by defining state/action spaces, reward functions, and generating initial code. The RL controller developed with DTE assistance outperformed traditional methods in simulation. Similarly, Tang et al. \cite{tang2024large} showcased LLM (GPT-4) assistance in designing arterial green wave policies. The LLM successfully parsed network files, calculated required signal offsets based on user inputs (distances, speed, cycle time), aided in policy evaluation via spatio-temporal diagrams, and helped adapt simulation code, improving average vehicle speeds in simulations. These studies highlight the utility of LLMs in streamlining complex engineering workflows and translating human expertise into operational parameters.

As research shifted towards LLMs as direct decision-makers, their function can often be abstractly represented as a policy $\pi_{\text{LLM}}$ selecting a control action $a_t$ based on the current traffic state $s_t$, embedded knowledge $K$ (e.g., traffic rules), and a guiding prompt structure $P$:
\begin{equation}
    a_t = \pi_{\text{LLM}}(s_t, K, P) \label{eq:llm_policy}
\end{equation}
Building upon assistive roles, researchers investigated LLMs as direct controllers for Adaptive Traffic Signal Control (ATCS). Movahedi and Choi \cite{movahedi2024crossroads} framed ATCS as a reasoning problem amenable to LLMs. They developed two GPT-3.5-Turbo \cite{openai_gpt35turbo_2025} based controllers: a Zero-Shot Chain of Thought (ZS-CoT) controller relying solely on prompted reasoning, and a Generally Capable Agent (GCA) employing an Actor-Critic structure where an LLM actor made decisions and an LLM critic evaluated them to update a text-based knowledge base. The GCA's ability to learn from feedback resulted in significantly better performance (reduced halts, increased speed, lower delay) compared to conventional controllers and the ZS-CoT approach, underscoring the value of incorporating learning mechanisms, although the need for safety overrides for potentially unsafe LLM outputs was noted.

Further exploring LLMs as direct TSC agents, Lai et al. \cite{lai2023large} proposed LLMLight. This approach transforms real-time traffic data into structured prompts, enabling LLMs to select optimal signal phases using Chain-of-Thought (CoT) reasoning. This process can be formalized as generating a rationale $r_t$ which is then used to extract the action $a_t$:
\begin{equation}
    \begin{aligned}
        r_t &= \text{CoTLLM}(s_t, P) \\
        a_t &= \text{Extract}(r_t)
    \end{aligned} \label{eq:llmlight}
\end{equation}
Recognizing the limitations of general-purpose LLMs, they developed LightGPT, a specialized Llama-based model fine-tuned specifically for TSC through imitation learning and critic-guided refinement. LightGPT demonstrated superior effectiveness, scalability, and robustness compared to general LLMs and RL baselines across diverse datasets, adapting well to unseen networks and extreme traffic conditions while providing interpretable rationales.

Architectural innovations continue to shape LLM-based TSC. Wang et al. \cite{wang2024llm} introduced LA-Light, positioning the LLM as a central decision-maker enhanced by specialized perception and decision tools. Tested in challenging simulation scenarios (emergency vehicles, roadblocks, sensor failures), LA-Light significantly outperformed conventional methods, particularly excelling in handling rare events and prioritizing emergency vehicles without retraining. Its capability to infer conditions during sensor outages using inherent common sense reasoning was a notable advantage, though latency remained a practical concern. Masri et al. \cite{masri2025large} conceptualized LLM controllers within a 4D framework (Detect, Decide, Disseminate, Deploy). Using CoT prompting on simulated intersection data, their fine-tuned GPT-4o-mini model demonstrated high accuracy (83\%) in detecting and resolving traffic conflicts based on rules, showcasing strong decision-making capabilities. Tang et al. \cite{tang2024largeurban} proposed a comprehensive framework based on the ACP methodology, where the LLM acts as an intelligent interface facilitating human-computer interaction and supporting multiple operational modes (autonomous, feedback, human takeover) to ensure reliability and safety.

Addressing limitations like poor generalization and interpretability in prior TSC methods, Zhang et al. \cite{zhang2024ragtraffic} proposed RAGTraffic. This framework utilizes Retrieval-Augmented Generation (RAG) to enhance LLM decision-making by integrating historical and real-time data via an adaptive retrieval mechanism (TRIS) coupled with iterative CoT reasoning. RAGTraffic aims for more human-like, context-aware reasoning, explicitly mitigating LLM hallucination risks and improving adaptation to dynamic conditions, outperforming baseline RL methods in performance, adaptability, and decision transparency.

Hybrid approaches combining the strengths of LLMs and RL are also proving effective, creating synergistic systems. Pang et al. \cite{pang2024illm} introduced iLLM-TSC, where an RL agent makes an initial decision ($a_t^{\text{RL}}$), which is then reviewed and potentially refined into a final action ($a_t^{\text{final}}$) by an LLM agent using structured prompts encoding rules and context ($K$). This synergistic approach leverages RL's policy learning efficiency and the LLM's reasoning and generalization capabilities, particularly enhancing robustness against real-world complexities like degraded communication. The core mechanism can be formalized as:
\begin{equation}
    \begin{aligned}
        a_t^{\text{RL}} &= \pi_{\text{RL}}(s_t) \\
        a_t^{\text{final}} &= \text{LLMreview}(s_t, a_t^{\text{RL}}, K)
    \end{aligned} \label{eq:illm_tsc}
\end{equation}
Experiments showed iLLM-TSC significantly reduced vehicle waiting times (by 17.5\% vs. standard RL) under degraded communication and effectively managed rare events like emergency vehicle prioritization.

Finally, bridging the simulation-to-reality gap is critical. Da et al. \cite{da2023llm} addressed this using PromptGAT, augmenting the Grounded Action Transformation (GAT) framework. By prompting an LLM (GPT-4 \cite{achiam2023gpt}) with contextual information (weather, traffic state), they inferred dynamic parameters fused into GAT's forward model, significantly reducing the performance gap when transferring RL policies between simulators with different dynamics, enhancing the real-world applicability of learned policies.

\subsection{Human Mobility Pattern Analysis and Synthesis}

Understanding and predicting human mobility patterns is fundamental for urban planning, traffic management, public health response, and infrastructure development. However, traditional methods often struggle with the complexity of human behavior, data sparsity, privacy concerns associated with data collection, and capturing the influence of external events. Large Language Models (LLMs) also offer promising avenues to address these challenges by leveraging their capabilities in reasoning, contextual understanding, and data generation.

Several studies have explored LLMs for predicting human mobility, particularly next location prediction. Liu et al. \cite{liu2024nextlocllm} proposed {NextLocLLM}, directly integrating an LLM into the prediction architecture. Departing from traditional discrete location IDs, their model utilizes continuous spatial coordinates to better capture geographic relationships and improve generalization across different cities. A key innovation is the use of LLM-enhanced Point of Interest (POI) embeddings, allowing the model to understand the functional attributes of locations based on POI descriptions. By combining spatial, functional, and trajectory data within a partially fine-tuned LLM, NextLocLLM demonstrated superior performance in both supervised and zero-shot prediction tasks compared to existing methods. Addressing zero-shot prediction specifically, Feng et al. \cite{feng2024agentmove} introduced {AgentMove}, an agentic framework arguing that direct prompting underutilizes LLM potential. AgentMove decomposes the task using dedicated modules: a spatial-temporal memory capturing individual patterns, a world knowledge generator leveraging textual addresses and urban structure for exploration modeling, and a collective knowledge extractor using location graphs from multiple trajectories. By integrating these information sources through reasoning prompts, AgentMove significantly improved zero-shot accuracy over baselines. Chen et al. \cite{chen2025toward} focused on enhancing interpretability and performance in data-scarce scenarios. Their approach uses a multi-round continuous dialogue mechanism, translating historical travel data into natural language prompts. The LLM performs step-by-step reasoning, using activity prediction as an intermediate step and incorporating predictions from deep learning models for self-correction. An entropy-weighted TOPSIS method refines candidate locations. This dialogue-based method achieved performance comparable to deep learning models on full datasets and superior accuracy in few-shot settings, crucially providing interpretable explanations for its predictions.

Beyond prediction, LLMs are also being employed to tackle challenges related to disruptive events and data scarcity. Zhong et al. \cite{zhong2024hmp} developed the {HMP-LLM} framework specifically to predict human mobility during major disruptions like pandemics. Their method decomposes mobility time series and uses a two-stage prompting strategy: the first stage learns historical patterns, and the second incorporates external event data (e.g., COVID-19 statistics) to refine predictions based on the LLM's reasoning abilities. Evaluated on Beijing data during the COVID-19 pandemic, HMP-LLM significantly outperformed traditional baselines, showcasing LLMs' strength in modeling mobility under complex, intervention-heavy conditions. Addressing the high cost and privacy issues of traditional travel surveys, Bhandari et al. \cite{bhandari2024urban} investigated using LLMs to {synthesize realistic travel survey data}. They prompted various LLMs (Llama-2 \cite{touvron2023llama}, Gemini-Pro \cite{team2023gemini}, GPT-4 \cite{achiam2023gpt}) with demographic context to generate daily travel diaries. While base LLMs showed relevant knowledge, fine-tuning an open-source model (Llama-2 \cite{touvron2023llama}) on limited real survey data dramatically improved the realism of the synthetic data. The fine-tuned model effectively replicated aggregate travel patterns, trip transitions, and activity chains, outperforming base LLMs and a traditional agent-based model, positioning fine-tuned LLMs as a viable tool for augmenting mobility datasets.

\subsection{Trajectory Prediction for Road Users}

Accurate prediction of the future movements of road users, including vehicles and pedestrians, is paramount for the safety and efficiency of autonomous driving (AD) systems and advanced driver-assistance systems (ADAS). Predicting trajectories involves understanding complex interactions, intentions, and adherence to traffic rules within dynamic environments. Large Language Models (LLMs) are increasingly being explored in this domain to leverage their capabilities in sequence modeling, scene understanding, reasoning, and even generating human-like explanations for predicted actions.

Several approaches integrate LLMs into the trajectory prediction pipeline. Lan et al. \cite{lan2024traj} introduced {Traj-LLM}, which applies pre-trained LLMs (like GPT-2 \cite{radford2019language}) to vehicle trajectory prediction without requiring explicit prompt engineering. They achieve this by transforming agent states and scene features into LLM-compatible token sequences. The LLM models complex scene contexts and social interactions, enhanced by a novel Lane-Aware Probabilistic Learning module using a Mamba state-space model to mimic human focus on relevant lanes. Efficiently fine-tuned using LoRA, Traj-LLM outperformed state-of-the-art methods on the nuScenes dataset, particularly in few-shot learning scenarios, demonstrating LLM potential for robust prediction. Focusing specifically on challenging maneuvers, Peng et al. \cite{peng2025lc} developed {LC-LLM} to improve long-term prediction and interpretability for vehicle lane change intentions and trajectories. By framing the task as a language modeling problem, they fine-tuned an LLM (Llama-2-13b-chat \cite{touvron2023llama}) using natural language prompts describing the driving scenario. Incorporating Chain-of-Thought (CoT) reasoning and explanatory prompts allowed the model to generate not only predictions but also underlying rationales, significantly boosting transparency. LC-LLM showed substantial improvements in both intention classification and trajectory accuracy over baselines on the highD dataset.

LLMs are also being applied to pedestrian trajectory prediction and integrated decision-making frameworks. Chib and Singh \cite{chib2024lg} proposed {LG-Traj} to enhance pedestrian trajectory prediction by incorporating semantic understanding. Their method uses an LLM to generate descriptive motion cues (e.g., 'linear motion') from observed coordinates. These LLM-generated cues, alongside trajectory data and interaction modeling within a transformer architecture, led to improved performance on the ETH-UCY \cite{pellegrini2009youll} and SDD \cite{robicquet2016learning} benchmarks. Beyond direct prediction, LLMs are used to generate realistic motion data for simulation. Ramesh and Flohr \cite{ramesh2024walk} developed the {"Walk-the-Talk"} framework to generate realistic 3D pedestrian motion sequences directly from natural language descriptions (e.g., "pedestrian jaywalking while looking at phone"). Using a specialized dataset focused on AD-relevant actions and a text-to-motion-token translation approach, they significantly improved the realism of pedestrian behavior in simulations, crucial for testing AD safety, especially for rare events. Similarly, Yang et al. \cite{yang2025trajectory} created {Trajectory-LLM} to generate realistic and diverse *vehicle* trajectory data from textual descriptions of interactions (e.g., "Car A bypasses Car B"). Their two-stage process uses an LLM first to interpret text and map context into driving behaviors and logic, and then guides the LLM to generate specific waypoint-based trajectories. Augmenting real data with this synthetic data improved downstream prediction models. Furthermore, LLMs are being integrated into driving decision pipelines. Yildirim et al. \cite{yildirim2024highwayllm} developed {HighwayLLM}, where an LLM assists a high-level reinforcement learning agent in highway driving. The LLM predicts waypoints based on the current state, surrounding vehicles, and retrieved similar historical trajectories, also providing a natural language rationale, thereby refining motion plans and significantly reducing collisions compared to an RL-only baseline.

\subsection{Simulation and Scenario Generation}

The rigorous testing and validation of autonomous driving (AD) systems and traffic management strategies heavily rely on realistic and diverse simulation scenarios. However, the traditional manual creation of these scenarios is often a laborious, time-consuming process that struggles to encompass the vast range of potential conditions, particularly rare but critical "edge cases" crucial for safety assurance \cite{chang2024llmscenario, cai2025text2scenario}. 

LLMs are being employed to interpret natural language instructions, enabling more intuitive and flexible control over simulation environments. For instance, Tan et al. introduced {ProSim}, a closed-loop traffic simulation framework where an LLM (LLaMA3-8B \cite{grattafiori2024llama} fine-tuned with LoRA) interprets complex textual prompts describing desired agent actions, interactions, or overall scenario characteristics \cite{tanpromptable}. This allows users to generate specific, complex traffic situations on demand by translating natural language descriptions into agent policies within the simulation, maintaining high fidelity and realistic interactions. Similarly, Li et al. developed {ChatSUMO}, leveraging an LLM (Llama 3.1 \cite{grattafiori2024llama}) to allow users to generate, modify, and analyze SUMO traffic simulations using natural language commands \cite{li2024chatsumo}. ChatSUMO can parse requests, generate networks, customize simulation elements, run simulations, and provide analytical summaries, significantly lowering the barrier to entry for using sophisticated simulation tools and facilitating rapid prototyping for mobility planning.

Beyond controlling existing simulators, LLMs are being utilized to generate entire traffic scenes or specific test scenarios directly from textual descriptions. Ruan et al. introduced a text-to-traffic scene generation pipeline using GPT-4o \cite{ruan2024traffic}. Their framework analyzes text, retrieves road segments, plans agent behaviors, and renders scenes in CARLA without predefined paths, leading to greater scenario variability and demonstrably improving AD agent performance in safety benchmarks. Addressing the need for critical test cases, Chang et al. developed {LLMScenario}, which uses meticulous prompt engineering (structured tokens, Chain-of-Thoughts reasoning) and a feedback loop based on reality/rarity metrics to generate rare and complex driving scenarios using models like ChatGPT4 \cite{chang2024llmscenario}. Concurrently, Cai et al. focused on standardizing test scenarios with {Text2Scenario (T2S)} \cite{cai2025text2scenario}. This framework converts natural language descriptions into executable ASAM OpenScenario scripts using a robust prompt pipeline (GPT-4 with CoT and self-consistency) and a hierarchical repository, efficiently generating feasible scenarios that successfully uncovered safety violations in various AD stacks. Furthermore, {SeGPT} leverages ChatGPT with CoT and domain-specific templates to parse user queries and synthesize complex, parameterized scenarios, improving the robustness of prediction algorithms and tackling data scarcity \cite{10423819}.

LLMs are also proving valuable in generating specific simulation components or analyzing simulation outputs. Wang et al. utilized ChatGPT-4 to automate the generation of diverse warehouse layouts for evaluating Multi-Agent Path Finding (MAPF) algorithms \cite{wang2024llmsim}, demonstrating a significant speed-up compared to manual design, though noting current limitations in creativity and handling complex constraints. Focusing on agent behavior, Xia et al. developed {InteractTraj} for language-driven interactive traffic trajectory generation \cite{xia2024language}. Using a two-stage "code bridge" (GPT-4 translating text to interaction codes, then a decoder generating trajectories), it produces realistic multi-agent interactions like overtaking and yielding, outperforming baselines in realism and faithfulness to commands. The "Walk-the-Talk" framework \cite{ramesh2024walk}, while primarily for trajectory prediction, also contributes by enabling the generation of realistic pedestrian behaviors from text, further enriching simulation realism. Complementing generation, Lu et al. introduced {DIAVIO}, an LLM-empowered framework (using fine-tuned Llama2 \cite{touvron2023llama}) to automate the diagnosis of safety violations found in AD simulations \cite{lu2024diavio}. By learning from real-world accident data via a crash-specific DSL, it accurately determines liability and classifies crash types, drastically reducing manual diagnosis time.

Expanding the scope beyond microscopic simulation, Zhang et al. introduced {ChatTraffic} for Text-to-Traffic Generation (TTG) \cite{zhang2024chattraffic}. This novel approach generates city-level traffic states (speed, congestion) based \emph{only} on textual descriptions of time and events (e.g., holidays, road closures), using a latent diffusion model with BERT and a GCN-augmented U-Net. ChatTraffic effectively simulates the impact of specific conditions, particularly abnormal events, offering a flexible tool for macroscopic traffic analysis and planning.

\subsection{Trip Planning and Navigation}

Trip planning and navigation are fundamental aspects of mobility, enabling users to efficiently reach their destinations while considering various constraints and preferences. LLMs offer promising approaches to enhance trip planning through improved natural language understanding, personalization, and context awareness.

Pio et al. \cite{pio2024framework} presented an interactive framework for intelligent trip planning that prominently features Large Language Models to enhance user experience and optimize itineraries. In their system, the LLM acts as the core component for natural language understanding, interpreting user requests detailing start/end points (or types of amenities), preferred transportation mode, and maximum time limits. The LLM extracts these parameters and translates the user's conversational input into structured Cypher queries executed against a Neo4j graph database containing OpenStreetMap data. This enables the generation of shortest paths incorporating relevant Points of Interest (POIs) and includes a mechanism to intelligently prune POIs if the cumulative travel and visit time exceeds the specified constraints, thereby improving personal mobility through user-friendly, customized, and time-aware travel planning, particularly beneficial for tourism.

Similarly, Tang et al. (\cite{tang2024synergizing}) developed a system called {ITINERA}. This innovative approach combines the natural language understanding capabilities of LLMs with traditional spatial optimization algorithms to create personalized urban itineraries that balance user preferences with spatial coherence. By decomposing complex user requests, retrieving relevant points of interest, and employing cluster-aware spatial optimization through hierarchical traveling salesman problem solutions, these integrated systems can generate routes that minimize travel distance while maximizing alignment with user needs. The successful integration of LLMs with spatial reasoning demonstrates significant potential for enhancing transportation systems, particularly in areas such as personalized route planning, traffic flow optimization, and the development of more intuitive, natural language interfaces for mobility applications—ultimately contributing to both improved user experiences and more efficient transportation networks.

Kuftinova et al. \cite{kuftinova2024large} explored LLMs for broader {suburban transport data management}. They discussed using LLMs to manage digital infrastructure data (digital twins), automate planning processes, generate route descriptions or traffic announcements, and analyze GPS data to understand movement patterns. They also proposed using LLMs within mathematical models (mixture distributions) for forecasting heterogeneous transport demand.

Leong et al. \cite{leong2024metroberta} developed {MetRoBERTa}, an LLM (RoBERTa-based) fine-tuned specifically on transit agency Customer Relationship Management (CRM) text data to classify unstructured customer feedback into relevant transit topics (e.g., delays, cleanliness, customer service, unsafe driving, security). MetRoBERTa significantly outperformed traditional TF-IDF based ML models due to its superior contextual understanding. They demonstrated an analytical pipeline combining topic classification with sentiment analysis, entity extraction (mode, route, location), and normalization using ridership data, enabling actionable insights like geospatial visualizations of complaints and trend tracking. This showcases LLMs' power in extracting structured insights from unstructured user feedback to improve mobility services.

\begin{table*}[!htbp]
    \footnotesize 
    \caption{Qualitative Summary of LLM Contributions to Mobility Enhancement}
    \label{tab:llm_mobility_summary_qualitative}

    \renewcommand{\arraystretch}{1.3}
    \setlength{\tabcolsep}{5pt} 

    \begin{tabularx}{\textwidth}{@{}
    >{\RaggedRight\arraybackslash}p{0.15\textwidth}  
    >{\RaggedRight\arraybackslash}X                 
    >{\RaggedRight\arraybackslash}X                 
    >{\RaggedRight\arraybackslash}p{0.15\textwidth}  
@{}}
    \toprule
    \textbf{Application Area} & \textbf{LLM Contribution Highlight} & \textbf{Impact / Enhancement} & \textbf{Representative Citations} \\
    \midrule

    \textbf{Traffic Flow Prediction and Forecasting}
    &
    Adapting LLM architectures (BERT, GPT-2, Llama) for time-series using specialized embeddings, partial freezing, text reprogramming, or semantic features; Enabling zero/few-shot forecasting.
    &
    Improves forecasting accuracy (esp. long-range, few-shot, zero-shot), handles complex spatio-temporal patterns, incorporates contextual factors (events, weather), enhances resource allocation and congestion management.
    &
    \cite{JIN2021115738}, \cite{liu2024spatial}, \cite{ren2024tpllm}, \cite{li2024urbangpt}, \cite{huang2024enhancing}, \cite{jin2023time}, \cite{chang2023llm4ts}, \cite{rasul2023lag}
    \\ \midrule

    \textbf{Traffic Data Analysis and Decision Support}
    &
    Providing natural language interfaces for querying complex mobility databases (SQL generation, ontology construction); Orchestrating specialized tools/models (TFMs) for analysis; Automating ontology creation.
    &
    Democratizes data access for non-experts, enables interactive decision support via conversational agents, automates knowledge base construction, integrates diverse analysis tools efficiently.
    &
    \cite{li2023can} , \cite{cai2022star}, \cite{padoan2024mobility}, \cite{yang2025independent}, \cite{tupayachi2024towards}, \cite{zhang2024semantic}, \cite{zhang2024trafficgpt}, \cite{da2024open}, \cite{wang2024leveraging}, \cite{jonnala2025exploring}
    \\ \midrule

    \textbf{Traffic Signal Control and Optimization}
    &
    Acting as reasoning agents (DTEs, controllers) to design, evaluate, or directly control signals using NL instructions/prompts (CoT, RAG); Handling rare events; Refining RL agent decisions.
    &
    Automates/assists complex signal timing design, improves traffic flow (reduces delay/stops), enables adaptive control responsive to real-time conditions and rare events (EMVs, incidents), enhances interpretability.
    &
    \cite{dai2024large}, \cite{tang2024large}, \cite{movahedi2024crossroads}, \cite{lai2023large}, \cite{wang2024llm}, \cite{masri2025large}, \cite{zhang2024ragtraffic}, \cite{pang2024illm}, \cite{da2023llm}
    \\ \midrule

    \textbf{Human Mobility Pattern Analysis and Synthesis}
    &
    Performing interpretable next location prediction via dialogue/prompts; Synthesizing realistic travel diaries; Extracting semantic understanding (intentions/preferences) from mobility data.
    &
    Improves accuracy/interpretability of mobility predictions (esp. few-shot), enables generation of realistic synthetic data for planning, provides deeper understanding of travel behavior semantics.
    &
    \cite{chen2025toward}, \cite{qin2025lingotrip}, \cite{gong2024mobility}, \cite{wang2024ai}, \cite{ge2025llm}
    \\ \midrule

    \textbf{Trajectory Prediction for Road Users}
    &
    Applying LLMs for vehicle/pedestrian prediction by tokenizing states/features, using NL prompts for interaction/intention, or generating realistic motion/trajectory data from text descriptions.
    &
    Improves accuracy and interpretability of trajectory predictions (esp. long-term, interactive), enhances AV safety by anticipating agent movements, enables generation of realistic training/simulation data from descriptions.
    &
    \cite{lan2024traj}, \cite{peng2025lc}, \cite{chib2024lg}, \cite{ramesh2024walk}, \cite{yang2025trajectory}, \cite{yildirim2024highwayllm}
    \\ \midrule

    \textbf{Simulation and Scenario Generation}
    &
    Generating diverse/critical/realistic traffic scenarios (CARLA, SUMO) or simulation inputs (layouts) from NL descriptions; Automating scenario scripting (OpenScenario) or violation diagnosis.
    &
    Accelerates/diversifies AV testing, enables generation of rare/critical scenarios via text, automates tedious scenario creation/diagnosis, improves simulation fidelity and realism for training/validation.
    &
    \cite{cai2025text2scenario}, \cite{tanpromptable}, \cite{li2024chatsumo}, \cite{ruan2024traffic}, \cite{lu2024diavio}, \cite{zhang2024chattraffic}, \cite{wang2024llmsim}, \cite{xia2024language}, \cite{ramesh2024walk}
    \\ \midrule

    \textbf{Trip Planning and Navigation}
    &
    Acting as natural language interfaces for personalized trip planning, translating requests into database queries or optimization inputs; Generating route descriptions/advice; Classifying user feedback for insights.
    &
    Enhances user experience, enables personalized/context-aware itineraries, simplifies access to complex routing/POI data, provides actionable insights from user feedback for service improvement.
    &
    \cite{pio2024framework}, \cite{tang2024synergizing}, \cite{kuftinova2024large}, \cite{leong2024metroberta}, \cite{wang2024leveraging}
    \\ \midrule

    \textbf{Transport Mode Choice Prediction}
    &
    Performing zero/few-shot mode choice prediction using prompt engineering or textual data representations; Capturing underlying semantics and preferences; Predicting choices during disruptions.
    &
    Improves prediction accuracy (esp. few-shot), enhances model interpretability, captures contextual/preference factors better than traditional models, aids understanding passenger responses to delays.
    &
    \cite{qin2025lingotrip}, \cite{gong2024mobility}, \cite{wang2024ai}, \cite{ge2025llm}, \cite{yang2024applying}
    \\ \midrule

    \textbf{Parking Planning and Management}
    &
    Simulating driver parking search behavior (personas); Interpreting complex parking signs (lightweight LLMs); Providing conversational parking assistance; Evaluating parking facilities (agent framework).
    &
    Aids research via realistic behavior simulation, improves driver understanding of parking rules (on-device potential), enhances in-car assistance safety/relevance, supports data-driven parking infrastructure planning.
    &
    \cite{fulman5018154utilizing}, \cite{OchubaInterpretingPS}, \cite{rony2023carexpert}, \cite{jin2024large}
    \\
    \bottomrule
    \end{tabularx}
\end{table*}

\subsection{Transport Mode Choice Prediction}

Understanding and predicting how individuals choose their mode of transport is vital for transportation planning, traffic management, and policy design. Traditional discrete choice models often rely on numerical data and utility maximization theory, which may struggle to capture the full complexity of human decision-making processes, heterogeneity among travelers, and the influence of contextual factors. Recent studies explore the use of Large Language Models (LLMs) to address these limitations.

Several studies investigate LLMs for direct mode choice prediction, often employing prompt engineering to guide the models. Mo et al. (2023) demonstrated that GPT-3.5, using zero-shot prompting with carefully engineered prompts including task descriptions, travel attributes, individual characteristics, and domain knowledge guides, could achieve prediction accuracy competitive with standard supervised methods (multinomial logit, random forest) on the Swissmetro dataset without any training data \cite{mo2023large}. However, they noted reliability challenges like reasoning errors. Liu et al. (2024), using the same dataset, found significant behavioral misalignment in zero-shot LLMs (GPT-3.5/GPT-4 \cite{achiam2023gpt}) but showed that alignment techniques like few-shot learning or a novel "travel behavior persona loading" method significantly improved accuracy, even surpassing traditional models when domain knowledge was incorporated \cite{liu2024can}. Zhai et al. (2024) proposed transforming input variables into textual descriptions and using demonstrations (including panel data showing past user choices) with LLMs (LLaMA3, Gemma) to predict mode choice on London and Swiss datasets, outperforming both traditional and deep learning models while offering superior interpretability through textual explanations \cite{zhai2024enhancing}.

LLMs are also applied to predict travel choices in specific contexts, such as during service disruptions or for predicting future trips. Chen et al. (2024) developed the DelayPTC-LLM framework to predict metro passenger choices (wait vs. abandon trip) during delays, finding that LLMs with tailored prompt engineering (including Chain-of-Thought) significantly outperformed traditional methods on Shenzhen Metro data, providing actionable insights for managing disruptions \cite{chen2024delayptc}. Qin et al. (2025) proposed {LingoTrip}, using zero-shot In-Context Learning with sophisticated spatiotemporal prompts (incorporating long/mid/short-term mobility patterns and real-world location names) to predict the next public transport destination from Hong Kong MTR data, showing superior accuracy, especially in zero/few-shot scenarios, and interpretability \cite{qin2025lingotrip}. Gong et al. (2024) introduced {Mobility-LLM}, a unified framework using reprogrammed check-in sequences, a Visiting Intention Memory Network, and Human Travel Preference Prompts to allow LLMs (fine-tuned with LoRA) to excel at next location prediction, trajectory user linking, and time prediction across multiple datasets, showing strong few-shot learning capabilities \cite{gong2024mobility}.

Beyond direct prediction, LLMs are used to simulate more complex travel behaviors and understand underlying patterns. Wang et al. (2024) introduced {LLMTraveler}, an LLM-agent framework with memory and personality traits, which successfully replicated complex day-to-day route-switching patterns in network congestion games that traditional models often miss, while demonstrating sample efficiency and interpretability \cite{wang2024ai}. Ge et al. (2025) used a fine-tuned LLM (Llama 3) within a cognitive modeling architecture to generate personalized travel plan suggestions, showing superior performance in reproducing aggregate travel flows and generating more diverse travel strategies compared to traditional cognitive models \cite{ge2025llm}. Li et al. (2025) leveraged semantic embeddings from BERT (treating multimodal trips as textual sequences) to identify 35 distinct multimodal travel patterns in the Beijing Geolife dataset through clustering, offering finer-grained insights than traditional methods \cite{li2025understanding}. Complementing this, Li et al. (2024) developed {MobAgent}, an LLM agent framework that first understands mobility reasons from survey data and then uses recursive reasoning with individual profiles to generate realistic, personalized daily travel diaries, outperforming baselines in statistical accuracy \cite{li2024more}.

Yang et al. \cite{yang2024applying} specifically explored Masked Language Models (MLMs) like BERT by representing trip records as "sentences". MLM-derived features captured contextual factors better than traditional distance/time-based models (which had low accuracy around 30\%) and improved mode classification F1 scores when integrated into supervised models, demonstrating the potential of LLMs for nuanced behavioral understanding from textual representations of travel \cite{yang2024applying}.

Collectively, these studies indicate that LLMs, through various approaches including sophisticated prompting, agent-based modeling, fine-tuning, and semantic analysis, offer powerful new tools for predicting transport mode choices, understanding complex travel behaviors, and simulating traveler decisions with greater realism, heterogeneity, and interpretability than many conventional methods. While challenges like reliability and behavioral alignment remain active research areas, LLMs hold significant promise for enhancing travel demand modeling and mobility analysis.

\subsection{Parking Planning and Management}

Parking availability, regulations, and search behavior significantly impact urban mobility, traffic flow, and driver experience. Large Language Models (LLMs) are being explored to address various challenges in this domain, from understanding driver decisions to interpreting complex rules and aiding in infrastructure planning.

One area of investigation is using LLMs to simulate driver parking search behavior. Fulman et al. utilized GPT-4o mini to act as different driver "Personas" in static choice experiments and a dynamic serious game \cite{fulman5018154utilizing}. The LLM replicated established human patterns, such as preferences for lower costs and times, showed rational trade-offs, and emulated the influence of sociodemographic traits like income and age. It also demonstrated bounded rationality and risk sensitivity akin to human decision-making under uncertainty. This suggests LLMs could serve as valuable tools for preliminary analyses and hypothesis generation in parking behavior research, complementing traditional methods, although limitations like bias and spatial reasoning constraints exist.

LLMs are also being applied to help drivers navigate complex parking regulations in real-time. Ochuba explored fine-tuning lightweight LLMs (\(<3\)B parameters) like Danube, Gemma, and Stable LM for on-device interpretation of parking signs \cite{OchubaInterpretingPS}. Using Parameter-Efficient Fine-Tuning (PEFT) on a curated dataset, these compact models achieved high accuracy (\>90\%) in classifying parking permissions based on sign transcriptions and context (time/day), running efficiently on consumer hardware. This highlights the potential for smaller, privacy-preserving LLMs in driver assistance systems to reduce parking violations. Similarly, Rony et al. developed {CarExpert}, a retrieval-augmented conversational system using GPT-3.5-turbo for in-car environments \cite{rony2023carexpert}. By incorporating control mechanisms like input filtering and grounding answers in domain-specific documents (e.g., owner's manuals), CarExpert can safely answer driver queries about vehicle features, including parking assistance systems (like Park Assist operation or space requirements), thereby aiding drivers and potentially preventing damage.

Furthermore, LLMs are being considered for strategic parking infrastructure planning, especially considering the advent of Autonomous Vehicles (AVs). Jin and Ma proposed using an {LLM as a parking planning agent} (using GPT-4 \cite{achiam2023gpt}) to evaluate parking facilities during the transition to mixed AV/Human-Driven Vehicle (HDV) traffic \cite{jin2024large}. Their framework integrates indicator selection, weighting (using AHP), data analysis, and visualization, guided by structured prompts to mitigate LLM risks. Tested on real-world city data and SUMO simulations, the framework demonstrated high comprehensiveness and consistency in selecting relevant indicators (parking demand, land value, etc.) and reasonable success in evaluating parking adequacy under varying AV penetration rates. This showcases LLMs' potential utility in complex urban planning tasks, helping cities adapt parking infrastructure for future mobility shifts.

\section{LLM Applications for Roadway Safety Enhancement}

Beyond improving mobility efficiency, LLMs are being applied to various aspects of roadway safety, from analyzing past incidents to predicting future risks and enhancing vehicle and pedestrian safety systems.

\subsection{Crash Data Analysis and Reporting}
\label{subsec:crash_analysis}

Understanding the contributing factors and patterns within traffic crash data is fundamental to developing effective roadway safety countermeasures. Traditionally, this involves analyzing structured data fields and manually reviewing often extensive and unstructured narrative portions of crash reports—a laborious process. Large Language Models (LLMs) present transformative potential in this domain, offering capabilities to automate the extraction of nuanced information, identify hidden patterns, and even predict outcomes from diverse data sources, including textual narratives and multimodal inputs.

A significant challenge lies in leveraging the rich detail contained within unstructured text. LLMs excel at processing natural language, enabling deeper analysis of crash narratives. For instance, Arteaga and Park \cite{arteaga2025large} demonstrated an LLM framework (testing ChatGPT, Flan-UL2, Llama-2) capable of analyzing textual crash reports to uncover underreported factors, such as alcohol involvement often missed or miscoded in structured fields. Their zero-shot approach, particularly using Flan-UL2 \cite{chung2022ul2} with implicit prompting, achieved high accuracy (F1-score of 0.96) compared to manual annotation, showcasing the potential for LLMs to significantly enhance data quality and analytical efficiency \cite{arteaga2025large}. However, consistency across different models remains a consideration. Mumtarin et al. \cite{mumtarin2023large} compared ChatGPT, Gemini, and GPT-4 on narrative analysis tasks, finding high agreement for simple binary questions but lower similarity for more complex interpretations like collision manner, suggesting caution and potential benefits of cross-validating results with multiple LLMs \cite{mumtarin2023large}.

Beyond analyzing existing narratives, LLMs are being used to enhance the analysis of structured data. Zhen et al. \cite{zhen2024leveraging} explored converting tabular crash data into detailed textual narratives. These narratives were then processed by LLMs (GPT-3.5, LLaMA3) using Chain-of-Thought (CoT) prompting to infer crash severity. The CoT approach not only improved performance but also provided interpretability by generating step-by-step reasoning that considered various factors (environmental, driver, vehicle, road geometry), addressing the 'black-box' nature of some traditional models. Both Prompt Engineering (PE) and CoT were found crucial for optimizing performance and handling sensitive labels \cite{zhen2024leveraging}. Similarly, Grigorev et al. \cite{grigorev2025enhancing, grigorev2024integrating} developed a hybrid machine learning approach where tabular report data, including narratives, was converted to full text. Features extracted by various LLMs (BERT \cite{devlin-etal-2019-bert}, XLNet \cite{yang2019xlnet}, RoBERTa \cite{liu2019roberta}, ALBERT \cite{lan2019albert}) were combined with baseline report features, significantly improving severity classification accuracy (e.g., F1 score boosted from 0.82 to 0.89 using BERT+XGBoost) compared to using baseline features alone, highlighting the value LLMs add in extracting contextual information \cite{grigorev2025enhancing, grigorev2024integrating}.

Further advancing predictive capabilities, Fan et al. \cite{fan2024learning} treated crash analysis as a language reasoning task. They created the large-scale CrashEvent dataset by textualizing crash reports with associated infrastructure and environmental data. Their fine-tuned Llama-2 model, CrashLLM, directly predicted crash outcomes (type, severity, injuries) from this textual context, substantially outperforming traditional machine learning classifiers (average F1 score improved from 34.9\% to 53.8\%). Critically, CrashLLM enables "what-if" causal analyses, allowing researchers to simulate the impact of specific factors (like alcohol use or icy roads) on crash distributions, offering powerful insights for intervention design and policy-making \cite{fan2024learning}.

LLMs are also being applied to analyze data from non-traditional sources and in real-time. Jaradat et al. \cite{jaradat2024multitask, jaradat4867993multi} employed a fine-tuned LLM (GPT-2 \cite{radford2019language}) within a multitask learning (MTL) framework to analyze Road Traffic Collisions (RTCs) using real-time Twitter data. Their system simultaneously performed classification (occurrence, severity, sentiment) and information retrieval (location, factors, injuries) with high accuracy (85\% average classification) and strong retrieval scores (ROUGE-I 0.78), enabling rapid incident monitoring and analysis using crowdsourced information \cite{jaradat2024multitask, jaradat4867993multi}.

Finally, there is growing interest in leveraging LLMs for automated crash reporting. Kuftinova et al. \cite{kuftinova2024large} discussed the potential of using LLMs to process multimodal data from smartphones (images, audio, GPS, text) to automatically generate comprehensive crash reports. While developed for construction, the AutoRepo framework by Pu et al. \cite{pu2024autorepo}, which uses unmanned vehicles and multimodal LLMs (MiniGPT4-7B) with techniques like LoRA/QLoRA fine-tuning for automated inspection reporting, demonstrates technologies adaptable to roadway contexts \cite{pu2024autorepo}. Furthermore, Wang et al. \cite{wang2023accidentgpt} introduced AccidentGPT, a sophisticated multimodal model integrating V2X perception data (from multiple cameras and vehicles/roadside units) with GPT-4V's reasoning capabilities for comprehensive accident analysis, prevention, and automated report generation suitable for traffic management agencies \cite{wang2023accidentgpt}.

\subsection{Driver Behavior Analysis and Risk Assessment}
\label{subsec:driver_behavior}

Understanding, predicting, and influencing driver behavior are paramount for enhancing roadway safety and facilitating the acceptance of autonomous vehicles (AVs). Large Language Models (LLMs) and Vision-Language Models (VLMs) are increasingly being utilized to analyze driver actions, assess risks associated with specific behaviors like distraction or fatigue, model nuanced human driving styles, and even detect malicious behaviors within connected vehicle networks.

One key application area is the direct analysis of driver state using sensor data. Zhang et al. \cite{zhang2024integrating} developed the Distracted Driving Language Model (DDLM), which integrates a VLM (LLaVA \cite{liu2023visual}) with whole-body pose estimation (HRNet). By analyzing driver postures, DDLM classifies various distracted behaviors (e.g., texting, eating) and assesses the associated risk level with high accuracy in both zero-shot and few-shot settings, outperforming baseline models and offering interpretable outputs suitable for driver monitoring systems \cite{zhang2024integrating}. Expanding on multimodal analysis, Takato et al. \cite{takato2024multi} developed a Multi-Frame VLM fine-tuned on synchronized video from road-facing and driver-facing cameras. Using a specialized visual instruction tuning dataset covering risky scenarios and driver states (distraction, fatigue), their model accurately interprets complex events and generates explanatory coaching advice, particularly aimed at improving safety in commercial fleets through automated, detailed behavior analysis \cite{takato2024multi}.

Beyond analyzing current behavior, LLMs are instrumental in modeling human driving characteristics to improve AV performance and interaction. Yang et al. \cite{yang2024driving} proposed a multi-alignment framework to imbue LLM-powered driver agents with specific human driving styles (e.g., 'cautious', 'risky'). Using a novel dataset capturing human driving rationale in natural language, along with an LLM 'Coach Agent' providing feedback, they successfully created agents exhibiting distinct styles in simulation, impacting safety metrics and human perception, thereby enhancing predictability and trust in AVs \cite{yang2024driving}. Similarly, Jin et al. \cite{jin2024surrealdriver} introduced the SurrealDriver framework, which aligns LLM agents with human cognitive processes. They collected a unique 'Driving-Thinking Dataset' of drivers' self-reported reasoning, using it to provide chain-of-thought demonstrations and inform driving guidelines within the LLM agent via a CoachAgent. This integration significantly reduced collision rates (by over 80\%) and increased perceived human-likeness (by 50\%) in simulations, highlighting the benefit of incorporating human cognitive patterns \cite{jin2024surrealdriver}.

LLMs are also being integrated as high-level reasoning components within AV control systems. Sha et al. \cite{sha2023languagempc} presented LanguageMPC, where an LLM acts as the high-level 'brain', performing attention allocation, situation awareness, and issuing textual action guidance (e.g., "slightly left slower") which is translated into parameters for a low-level Model Predictive Control (MPC) backend. This hierarchical approach improved performance and allowed for text-modulated driving styles \cite{sha2023languagempc}. Yildirim et al. \cite{yildirim2024highwayllm} also employed an LLM in their HighwayLLM framework to provide natural language reasoning alongside waypoint predictions for lane changes, enhancing the interpretability of AV decisions \cite{yildirim2024highwayllm}.

Enhancing the situational context available to LLMs is another avenue for improvement. Luo et al. \cite{luo2025senserag} proposed SenseRAG, a Retrieval-Augmented Generation framework that integrates real-time multimodal sensor data (camera, LiDAR, traffic signals, weather) into a language-accessible knowledge base. By proactively identifying uncertainties and querying this knowledge base using chain-of-thought prompting, the LLM can perform more accurate, context-rich reasoning, significantly reducing displacement errors in autonomous driving tasks \cite{luo2025senserag}.

Furthermore, LLMs contribute to safety by improving the simulation environments used for AV testing. Nguyen et al. \cite{nguyen2024text} introduced Text-to-Drive (T2D), a framework where an LLM first generates diverse textual descriptions of driving behaviors for background traffic and then generates the necessary components (state abstractions, reward functions) to train reinforcement learning agents that synthesize these behaviors realistically in simulation. This allows for the automated creation of varied and controllable scenarios, leading to more comprehensive AV testing \cite{nguyen2024text}.

Finally, LLMs are being applied to bolster security and safety in Connected Autonomous Vehicle (CAV) environments. Hu et al. \cite{hu2025llm} proposed using fine-tuned LLMs (via methods like Adapter/LoRA or prompt tuning) to detect malicious misbehavior, such as diffusion model-generated fake traffic signs or forged vehicle motion data broadcasts. The LLMs serve as verification layers, processing visual data or analyzing plausibility checks, and significantly outperform traditional methods in detecting these specific threats, thereby mitigating risks unique to connected mobility \cite{hu2025llm}.

\subsection{Pedestrian Safety and Behavior Modeling}
\label{subsec:pedestrian_safety}

Pedestrians represent a particularly vulnerable group of road users, and understanding, predicting, and accommodating their behavior is critical for preventing accidents and designing safer transportation systems. Large Language Models (LLMs) and Vision-Language Models (VLMs) are being applied in diverse ways to address challenges in pedestrian safety, ranging from analyzing past incidents to predicting future actions and assessing environmental risks.

One approach involves extracting insights from existing crash data. Das et al. \cite{das2023classifying} utilized Natural Language Processing (NLP) models, specifically BERT, to automatically classify pedestrian maneuver types directly from the unstructured textual narratives found in police crash reports. While achieving high accuracy (86\%) for binary classification of intentionality on Texas crash data, the model faced challenges with finer-grained multi-class maneuver classification due to data limitations, highlighting both the potential and the data dependency of using LLMs for analyzing historical incident descriptions \cite{das2023classifying}.

Predicting pedestrian intentions, especially regarding street crossings, is crucial for real-time safety applications like autonomous driving. Munir et al. \cite{munir2025pedestrian} proposed PedVLM, a multimodal VLM framework fine-tuned specifically for predicting pedestrian crossing intentions (crossing vs. not crossing) with enhanced explainability. By integrating RGB images, optical flow data, and tailored text prompts (using their PedPrompt dataset), PedVLM significantly outperformed baseline models and zero-shot GPT-4V approaches, demonstrating the value of domain-specific fine-tuning and multimodal inputs \cite{munir2025pedestrian}. Hussien et al. \cite{hussien2025rag} tackled explainable pedestrian crossing prediction using a combination of knowledge graphs (KGs), knowledge graph embeddings (KGEs), Bayesian inference, and an LLM-based Retrieval-Augmented Generation (RAG) component. Their KG-centric approach achieved state-of-the-art prediction performance on standard datasets (JAAD \cite{rasouli2017ICCVW}, PSI \cite{chen2021psi}), while the RAG module provided natural language explanations for the predictions, enhancing trust and transparency \cite{hussien2025rag}.

Real-time monitoring and risk assessment are also key areas. Abdelrahman et al. \cite{abdelrahman2024video, abdelrahman2025video} introduced Video-to-Text Pedestrian Monitoring (VTPM), a system designed to enhance safety while preserving privacy. Using computer vision to extract pedestrian movements, potential conflicts, and crossing violations (integrated with signal timing and weather data), the system employs LLMs (Phi-3 variants) to generate textual reports. A smaller LLM provides real-time, privacy-preserving summaries, while a larger, fine-tuned LLM enables interactive historical analysis of these text reports to identify safety trends. This approach significantly reduces data storage needs and mitigates privacy concerns associated with raw video footage \cite{abdelrahman2024video, abdelrahman2025video}. Focusing on vulnerable road users, Hwang et al. \cite{hwang2024safe} investigated using GPT-4V for interpretable risk assessment during street crossings, specifically to assist blind and low-vision pedestrians. Analyzing multiview egocentric images captured by a robot, they found that incorporating optical flow data alongside other visual knowledge (bounding boxes, segmentation masks) significantly improved GPT-4V's ability to accurately predict safety scores, highlighting the importance of temporal information and the potential of VLMs for nuanced safety-aware scene understanding \cite{hwang2024safe}.

Furthermore, LLMs contribute to pedestrian safety through improved simulation capabilities. Ramesh and Flohr \cite{ramesh2024walk} developed "Walk-the-Talk", an LLM-based framework capable of generating realistic, diverse 3D pedestrian motions directly from textual descriptions. This enables the creation of critical and potentially risky pedestrian behaviors (e.g., jaywalking, impaired walking) within simulation environments, facilitating more robust testing and validation of autonomous driving perception and prediction systems \cite{ramesh2024walk}.

LLMs are also being evaluated for assessing infrastructure relevant to pedestrian safety, though challenges remain. Shihab et al. \cite{shihab2024precise} studied the ONE-PEACE LLM for sidewalk detection via image segmentation. While the LLM showed strong accuracy under ideal conditions, its performance degraded significantly when subjected to real-world noise (Gaussian, Salt-and-Pepper, etc.), whereas a traditional ensemble learning model proved much more robust. This suggests that while LLMs hold promise for infrastructure assessment, their practical deployment may require hybridization or further research to improve resilience to noise \cite{shihab2024precise}.

\subsection{Traffic Rule Formalization and Compliance}
\label{subsec:rule_compliance}

Ensuring that autonomous vehicles (AVs) consistently understand and adhere to complex traffic rules and regulations, which are often expressed in ambiguous natural language, represents a significant challenge for roadway safety and deployment. Large Language Models (LLMs) are emerging as valuable tools to bridge the gap between human-readable laws and the precise logical representations or reasoning capabilities required for safe and compliant automated driving.

A key area of research involves translating traffic rules from natural language into formal logic systems suitable for machine verification and implementation. Manas et al. \cite{manas2024tr2mtl} introduced TR2MTL, a framework leveraging LLMs (including GPT-4 and GPT-3.5-turbo) to automatically convert traffic rules from sources like the German StVO \cite{stvo2021} and the Vienna Convention into Metric Temporal Logic (MTL) formulas. MTL is particularly useful as it can express the time-bound constraints often present in traffic regulations. By employing Chain-of-Thought (CoT) prompting and few-shot learning, TR2MTL achieved high translation accuracy (72.91\% with GPT-4+CoT), significantly outperforming simpler prompting methods. The resulting formalized MTL rules can be directly utilized for verifying AV behavior, constraining AV planning algorithms, and enabling real-time compliance monitoring, thus enhancing operational safety \cite{manas2024tr2mtl}.

Beyond static translation, LLMs are being integrated into dynamic decision-making frameworks to ensure real-time compliance. Cai et al. \cite{cai2024driving} proposed an interpretable framework where a Traffic Regulation Retrieval (TRR) Agent, based on Retrieval-Augmented Generation (RAG), dynamically fetches relevant rules, norms, and safety guidelines from comprehensive regional databases (including laws, manuals, and case precedents) pertinent to the vehicle's current context. This retrieved information is then passed to an LLM-powered Reasoning Agent, which interprets the complex semantics, distinguishes mandatory requirements from safety advice, and evaluates potential driving actions for both legal compliance and overall safety. This approach demonstrated adaptability across different regions and provides a pathway towards more reliable and interpretable AVs capable of navigating complex regulatory landscapes \cite{cai2024driving}.

LLMs are also being used to refine the decisions made by other control systems to improve compliance and safety in specific, complex scenarios. Yin et al. \cite{yin2024llm} developed a method where an LLM enhances the decisions of a primary Reinforcement Learning (RL) agent responsible for controlling vehicles in on-ramp merging areas. The RL agent makes initial action proposals based on the environment, which are then refined by the LLM using carefully designed prompts and chain-of-thought reasoning.
In reinforcement learning (RL) contexts, LLMs can refine the decisions of primary RL agents. The RL agent proposes an action $a_t$ at time $t$ based on its policy $\pi$:

\begin{equation}
    a_t = \pi(s_t)
\end{equation}
where $s_t$ is the current state. The LLM then evaluates and potentially refines $a_t$ using a prompt encoding the current scenario and relevant traffic rules, ensuring the final action $\tilde{a}_t$ adheres to both learned behaviors and formalized compliance constraints. This LLM-refinement process can be viewed as an additional policy layer:

\begin{equation}
    \tilde{a}_t = \text{LLM\_Refine}(a_t, \text{Rules}, s_t)
\end{equation}

This LLM-enhanced approach demonstrated improved traffic flow efficiency (average speed increased by approximately 14\%, and up to 22.1\% in dense traffic) compared to baseline RL methods, suggesting more effective and potentially safer merging maneuvers by mitigating erroneous decisions common in RL under novel conditions \cite{yin2024llm}.

\subsection{Near-Miss Detection}
\label{subsec:near_miss}

Near-miss incidents, defined as situations narrowly avoiding a crash, occur significantly more frequently than actual collisions and thus represent a rich source of data for proactive safety analysis. Detecting and analyzing these events can reveal hazardous locations, common risky behaviors, and system vulnerabilities before they lead to severe consequences. Large Language Models (LLMs) and Multimodal Large Language Models (MLLMs) are increasingly being employed, often in conjunction with computer vision (CV) techniques, to automate the detection and analysis of near-misses from video data.

A prominent approach involves a multi-stage framework combining CV models for initial event detection and LLMs/MLLMs for subsequent analysis. Jaradat et al. \cite{jaradat2024advanced, jaradat2025leveraging} developed systems using deep learning models (such as CNNs, ViT, or CNN+LSTM) to first process crowdsourced dashcam video footage. These models classify video segments or frames into categories like 'crash', 'non-crash', and 'near-miss', sometimes using specific classification probability thresholds (e.g., probabilities between 0.4 and 0.5 indicating a near-miss, while $>$ 0.5 indicates a crash \cite{jaradat2024advanced}). Once a near-miss or crash event is identified, an MLLM analyzes the relevant video portion. This analysis often involves generating detailed narrative descriptions of the event and extracting crucial safety-relevant features, such as the type of conflict, contributing environmental conditions, and actions of involved road users. This integrated methodology effectively leverages the high frequency of near-misses captured in video data to enable proactive safety interventions \cite{jaradat2024advanced, jaradat2025leveraging}.

Alternatively, MLLMs are being explored for end-to-end detection and analysis of safety-critical events directly from video. Abu Tami et al. \cite{abu2024using} utilized Gemini-Pro-Vision 1.5 to process driving video frames. They employed a structured, multi-stage question-answering (QA) prompting strategy to guide the MLLM through identifying potential risks, classifying the scene and involved agents, determining spatial relationships between entities, and even recommending appropriate driver actions in response to detected critical events (which include near-misses). Their evaluation on the DRAMA dataset \cite{malla2023drama} indicated that few-shot learning significantly enhanced the model's accuracy compared to zero-shot approaches, demonstrating the potential of guided MLLMs for comprehensive and actionable analysis of dynamic driving scenarios \cite{abu2024using}.

While not directly focused on traffic, methodologies from other domains involving LLMs for event prediction and causal analysis may offer transferable insights. For instance, Mudgal et al. \cite{mudgal2024crasheventllm} developed CrashEventLLM to predict computing system crashes and their causes from structured event sequences extracted from logs using LLMs (Llama2). The concepts of using LLMs for temporal pattern recognition in event sequences and performing abductive reasoning to identify potential causes could potentially be adapted for predicting traffic incidents, including near-misses, offering a direction for future research \cite{mudgal2024crasheventllm}.

\begin{table*}[!htbp]
    \footnotesize 
    \caption{Qualitative Summary of LLM Contributions to Roadway Safety Enhancement}
    \label{tab:llm_safety_summary_qualitative}

    \renewcommand{\arraystretch}{1.3}
    \setlength{\tabcolsep}{5pt} 

    \begin{tabularx}{\textwidth}{@{}
    >{\RaggedRight\arraybackslash}p{0.15\textwidth}  
    >{\RaggedRight\arraybackslash}X                 
    >{\RaggedRight\arraybackslash}X                 
    >{\RaggedRight\arraybackslash}p{0.15\textwidth}  
@{}}
    \toprule
    \textbf{Application Area} & \textbf{LLM Contribution Highlight} & \textbf{Impact / Enhancement} & \textbf{Representative Citations} \\
    \midrule

    \textbf{Crash Data Analysis \& Reporting}
    &
    Analyzing unstructured narratives/data to automatically extract factors, classify severity, identify underreporting, or generate reports; Processing social media for real-time updates.
    &
    Improves crash data quality and completeness significantly, speeds up analysis for quicker insights, enables data-driven countermeasure design, provides real-time incident awareness.
    &
    \cite{arteaga2025large}, \cite{mumtarin2023large}, \cite{zhen2024leveraging}, \cite{grigorev2025enhancing}, \cite{grigorev2024integrating}, \cite{fan2024learning}, \cite{jaradat2024multitask}, \cite{jaradat4867993multi}, \cite{wang2023accidentgpt}
    \\ \midrule

    \textbf{Driver Behavior Analysis \& Risk Assessment}
    &
    Interpreting multimodal data (vision, pose) for distraction/behavior classification; Generating human-like driving styles via reasoning/alignment; Guiding AV decision-making/planning; Detecting sophisticated CAV attacks.
    &
    Provides interpretable driver risk assessment, enables creation of more realistic and trustworthy AV agents, improves AV safety through better reasoning, enhances CAV security against novel threats.
    &
    \cite{zhang2024integrating}, \cite{takato2024multi}, \cite{yang2024driving}, \cite{jin2024surrealdriver}, \cite{sha2023languagempc}, \cite{yildirim2024highwayllm}, \cite{luo2025senserag}, \cite{nguyen2024text}, \cite{hu2025llm}
    \\ \midrule

    \textbf{Pedestrian Safety \& Behavior Modeling}
    &
    Classifying actions from text narratives, predicting intentions using VLMs, modeling behavior with explainability (KG/RAG), generating realistic motions from text descriptions, creating text summaries for privacy-preserving monitoring.
    &
    Automates analysis of pedestrian crash factors, enhances AV prediction capabilities, provides explainable safety models, enables better simulation testing, protects pedestrian privacy in monitoring systems.
    &
    \cite{das2023classifying}, \cite{munir2025pedestrian}, \cite{hussien2025rag}, \cite{abdelrahman2024video}, \cite{ramesh2024walk}, \cite{shihab2024precise}
    \\ \midrule

    \textbf{Traffic Rule Formalization \& Compliance}
    &
    Translating ambiguous natural language rules into precise, machine-readable formal logic (e.g., MTL); Retrieving and interpreting relevant regulations for AV decision-making (RAG).
    &
    Ensures AVs can understand and verifiably comply with complex regulations, enhancing safety and enabling consistent behavior; Supports adaptation to different regional rulesets.
    &
    \cite{manas2024tr2mtl}, \cite{cai2024driving}
    \\ \midrule

    \textbf{Near-Miss Detection}
    &
    Integrating CV and LLMs/MLLMs to automatically identify near-miss events from video footage and generate descriptive narratives for analysis.
    &
    Enables proactive safety interventions by leveraging often-unreported near-miss data, providing richer context and insights than traditional crash-only analysis.
    &
    \cite{jaradat2024advanced}, \cite{jaradat2025leveraging}, \cite{abu2024using}, \cite{mudgal2024crasheventllm} 
    \\ \midrule

    \textbf{Traffic Scene Understanding \& VQA}
    &
    Enabling natural language queries (VQA) about complex traffic scenes; Generating detailed captions/descriptions; Fusing multimodal inputs (vision, LiDAR, maps); Aligning model attention with human focus.
    &
    Allows intuitive interaction for scene analysis, improves AV/system understanding of multimodal contexts (incl. HD maps), enhances explainability, aids automated data annotation and monitoring.
    &
    \cite{cao2024maplm}, \cite{marcu2024lingoqa}, \cite{sun2024video}, \cite{keskar2025evaluating}, \cite{qasemi2023traffic}, \cite{onsu2025leveraging}, \cite{rekanar2024optimizing}, \cite{jain2024semantic}, \cite{rivera2025scenario}, \cite{dinh2024trafficvlm}
    \\
    \bottomrule
    \end{tabularx}
\end{table*}

\subsection{Traffic Scene Understanding and VQA}

A fundamental requirement for enhancing roadway safety and enabling advanced mobility solutions is the ability of AI systems to accurately perceive and comprehend complex, dynamic traffic environments. Visual Question Answering (VQA) and detailed scene understanding powered by Large Language Models (LLMs) and Vision-Language Models (VLMs) represent a significant step forward in this domain, moving beyond simple object detection towards nuanced interpretation and reasoning. These models are particularly adept at integrating visual information with textual queries and domain knowledge.

A primary challenge lies in the gap between general-purpose models and the specific demands of the autonomous driving domain. To address this, researchers are developing specialized datasets and benchmarks. Cao et al. \cite{cao2024maplm} introduced MAPLM, a large-scale, real-world dataset integrating panoramic 2D images, 3D LiDAR, and detailed HD map annotations. They also created the MAPLM-QA benchmark specifically for visual instruction tuning of VLMs for map and traffic scene understanding. Their baseline model, combining CLIP and an LLM, showed that instruction tuning on MAPLM-QA could yield meaningful representations, even outperforming zero-shot GPT-4V on certain tasks like lane counting, but highlighted remaining difficulties in achieving consistent frame-level accuracy \cite{cao2024maplm}. Similarly, Marcu et al. \cite{marcu2024lingoqa} presented LingoQA, a large dataset (419K QA pairs) focused on explainability in autonomous driving, enabling VLMs to answer free-form questions about driving actions (e.g., "Why did the vehicle slow down?"). Their work underscored current VLM limitations, particularly in temporal reasoning, where their baseline model achieved only $\approx$60\% accuracy compared to $\approx$97\% for humans, and introduced Lingo-Judge for efficient VQA evaluation \cite{marcu2024lingoqa}.

Beyond dataset creation, significant effort focuses on enhancing the reasoning capabilities of VLMs for traffic scenarios. Sun et al. \cite{sun2024video} tackled causal reasoning in traffic VideoQA. Recognizing that video frames alone are often insufficient, they proposed the TK-QC model, which classifies questions and utilizes a lightweight Traffic Knowledge Database (TKB) containing rules and event information. Causal questions are processed by a graph-based network (QK-AGA) incorporating retrieved knowledge alongside video features, leading to improved accuracy, especially on causal inference tasks within the TrafficQA dataset, while reducing computational overhead compared to monolithic models \cite{sun2024video}. Guo et al. \cite{guo2024cfmmc} addressed key gaps (dimension, scene, modality) hindering TrafficVQA performance. Their Coarse-Fine Multimodal Contrastive Alignment Network (CFMMC-Align) uses contrastive learning to align visual features and bridge the gap between visual context and textual answers, achieving state-of-the-art results on the SUTD-TrafficQA dataset \cite{Xu_2021_CVPR} and suggesting future integration with LLMs for even deeper semantic understanding \cite{guo2024cfmmc}. However, fine-grained spatial reasoning remains a hurdle, as demonstrated by Keskar et al. \cite{keskar2025evaluating}, who found that while NVIDIA's ViLA model excelled at high-level scene classification on MAPLM-QA, it struggled with precise tasks like lane counting and intersection recognition, pointing to the need for more robust VLM architectures and balanced datasets \cite{keskar2025evaluating}.

Researchers are also exploring methods to integrate external knowledge or pre-process inputs to improve VLM performance. Qasemi et al. \cite{qasemi2023traffic} developed TRIVIA, injecting traffic-domain knowledge by automatically generating descriptive captions from stationary camera footage using object detection and knowledge graphs (HANS framework). Fine-tuning VLMs like VIOLET on these video-caption pairs significantly boosted performance on the SUTD-TrafficQA benchmark \cite{qasemi2023traffic}. Onsu et al. \cite{onsu2025leveraging} enhanced LLaVA 1.5 \cite{liu2023improved} for traffic monitoring by integrating instance segmentation (YOLOv11s) to explicitly highlight vehicles and pedestrians before LLM processing. This pre-processing step, combined with fine-tuning, improved accuracy in recognizing vehicle location (84.3\%) and steering direction (76.4\%) \cite{onsu2025leveraging}. Rekanar et al. \cite{rekanar2024optimizing} improved VQA performance and interpretability by aligning machine attention with human attention. They applied a human-guided filter, based on features humans deem important (roads, lanes, signs, vehicles), to remove irrelevant object proposals before LXMERT processed the data, significantly improving results on nuImages \cite{rekanar2024optimizing}.

Comparative studies and evaluations further illuminate the capabilities and limitations of current models. Jain et al. \cite{jain2024semantic} evaluated Video-LLaVA and GPT-4 Vision on KITTI \cite{Geiger2012CVPR} and NuScenes \cite{nuscenes2019} using curated questions. GPT-4 generally provided more comprehensive answers, and integrating 3D tracking data improved interaction understanding but introduced hallucination risks, emphasizing the need for careful data fusion \cite{jain2024semantic}. Rivera et al. \cite{rivera2025scenario} benchmarked several LVLMs (including GPT-4 \cite{achiam2023gpt}, LLaVA \cite{liu2023visual}, CogVLM \cite{wang2024cogvlm}, ComposerHD \cite{zhang2023internlm}) for automatically categorizing urban traffic scenes across various detection and reasoning categories. While LVLMs had lower raw accuracy than fine-tuned CNNs on BDD100k, they exhibited higher F1 scores, suggesting better generalization without retraining, making them valuable for scalable dataset annotation and scene understanding \cite{rivera2025scenario}. Parthasarathy et al. \cite{parthasarathy2025glimpse}, addressing the LLVM-AD challenge on MAPLM, found that zero-shot Paligemma outperformed their fine-tuned CLIP \cite{radford2021learning} model for MCQ-based VQA under computational constraints, highlighting the critical trade-off between model size and performance for deployment on resource-constrained vehicle platforms \cite{parthasarathy2025glimpse}. Dinh et al. \cite{dinh2024trafficvlm} specifically developed TrafficVLM to generate fine-grained video captions of vehicle and pedestrian actions, offering richer semantic understanding than simple detection, crucial for incident analysis \cite{dinh2024trafficvlm}.

\section{Enabling Technologies and Frameworks}

Several cross-cutting technologies and architectural approaches are enabling and shaping the application of LLMs in transportation.

\subsection{V2X and Cooperative Driving Automation}

Vehicle-to-Everything (V2X) communication, encompassing vehicle-to-vehicle (V2V), vehicle-to-infrastructure (V2I), vehicle-to-pedestrian (V2P), and vehicle-to-network (V2N) interactions, forms a critical foundation for advanced driver assistance systems and Cooperative Driving Automation (CDA). The ability to share real-time information about vehicle states, intentions, and environmental conditions allows for enhanced situational awareness, improved traffic flow, and increased safety. Recently, Large Language Models (LLMs) have emerged as powerful tools within V2X and CDA frameworks, primarily due to their advanced capabilities in natural language understanding, reasoning, complex data interpretation, and decision-making support.

A significant challenge in V2X environments is processing and interpreting the vast amounts of heterogeneous data generated by diverse sources (vehicles, sensors, infrastructure). Wu et al. introduced the V2X-LLM framework specifically to tackle this issue within connected vehicle corridors \cite{wu2025v2x}. By integrating LLMs into an advanced data pipeline, their system performs real-time analysis through tasks such as Scenario Explanation, V2X Data Description, State Prediction, and Navigation Advisory. Field tests validated the framework's ability to enhance situational awareness and provide valuable decision support for traffic management, although challenges related to long-term prediction accuracy and processing latency were noted, suggesting needs for future model optimization \cite{wu2025v2x}. Complementing this, the survey by Panyam et al. highlights the role of LLMs in V2X data analysis, including pattern recognition and processing natural language reports, alongside real-time monitoring and anomaly detection \cite{panyam2025survey}. Kuftinova et al. also recognized the potential for LLMs to manage digital V2X infrastructure data effectively \cite{kuftinova2024large}.

Beyond data interpretation, LLMs are significantly enhancing cooperative perception. This relies on fusing information from multiple V2X sources (e.g., multi-camera feeds, sensor data from vehicles and infrastructure) to create a shared, panoramic understanding of the driving environment. Such data fusion can be mathematically modeled using principles like Bayesian inference to systematically combine evidence from different observations:
\begin{equation}
    P(S \mid O_1, O_2, \ldots, O_n) \propto P(S) \prod_{i=1}^{n} P(O_i \mid S)
    \label{eq:bayes_fusion} %
\end{equation}
where $P(S \mid O_1, \ldots, O_n)$ represents the estimated probability of the true environment state $S$ given observations $O_1, \ldots, O_n$ from $n$ different sources, incorporating prior knowledge $P(S)$. While various fusion techniques generate this comprehensive representation, the distinct advantage offered by LLMs and vision-language models (VLMs) lies in their capacity to subsequently process, interpret, and reason over this rich, fused information. This enables complex downstream tasks such as robust 3D object detection, accurate trajectory prediction, and high-level scene understanding, including accident analysis. 
For instance, Wang et al.'s AccidentGPT heavily relies on V2X communication to integrate panoramic multi-camera data from numerous vehicles and roadside units \cite{wang2023accidentgpt}. This collaboratively perceived environmental understanding is then fed into a GPT-based reasoning module for comprehensive accident analysis and prevention recommendations \cite{wang2023accidentgpt}. Similarly, You et al. developed V2X-VLM, an end-to-end framework utilizing VLMs \cite{you2024v2x}. By processing paired images from vehicle and infrastructure cameras alongside textual prompts, and leveraging V2X for enhanced, beyond-line-of-sight situational awareness, V2X-VLM demonstrated significantly improved performance and robustness, reducing collision rates on the DAIR-V2X dataset \cite{you2024v2x}.

LLMs are also being employed to orchestrate complex interactions in Cooperative Driving Automation (CDA). Fang et al. proposed CoDrivingLLM, a framework designed to manage CDA across all SAE J3216 levels \cite{sae_j3216_2021}, from simple status-sharing to prescriptive guidance \cite{fang2025towards}. In this system, an LLM acts as a central coordinator, receiving state and intent information from connected automated vehicles (CAVs), negotiating potential conflicts using traffic rules and time-to-conflict analysis, and making coordinated driving decisions, which are then pruned by safety checks. A memory module further allows the system to learn from past interactions. CoDrivingLLM significantly outperformed traditional optimization, rule-based, and reinforcement learning methods in simulated driving scenarios \cite{fang2025towards}.

Addressing the fundamental requirements of secure and efficient communication in V2X systems, Arshad et al. proposed the BlockLLM architecture \cite{arshad2025blockllm}. This framework integrates blockchain technology to ensure data integrity, non-repudiation, and trustless communication, while leveraging LLMs for adaptive decision-making and latency reduction in real-time data exchange. BlockLLM incorporates incentive and reputation mechanisms to encourage node participation and reliability. Simulations showed notable improvements, including an 18\% reduction in latency and a 12\% increase in throughput, indicating its potential as a robust and scalable solution for secure V2X \cite{arshad2025blockllm}. The need for robust architectures is echoed by Panyam et al., who discuss the necessity of hybrid edge-cloud setups to manage the inherent latency and computational demands of real-time V2X processing facilitated by LLMs \cite{panyam2025survey}.

\subsection{Domain Adaptation and Foundation Models}

While general-purpose Large Language Models (LLMs) offer powerful natural language understanding and generation capabilities, their direct application to specialized domains like transportation often falls short due to a lack of specific knowledge and the inability to effectively process domain-unique data types. To bridge this gap, significant research efforts are underway to adapt existing LLMs or construct bespoke foundation models tailored for roadway safety, traffic management, autonomous driving, and mobility enhancement. These approaches aim to imbue models with the necessary expertise and processing capabilities for transportation-specific challenges.

One primary strategy involves {fine-tuning pre-trained LLMs on domain-specific textual corpora}, where a pre-trained models' parameters ($\theta_{pre}$) are further trained on a domain-specific dataset to minimize a task-specific loss function ($L_{task}$). Zheng et al. \cite{zheng2023trafficsafetygpt} exemplified this by developing {TrafficSafetyGPT}. They fine-tuned the {LLaMA-7B \cite{touvron2023llama}} model using transportation safety knowledge meticulously extracted from official government guidelines, such as the NHTSA's Model Minimum Uniform Crash Criteria (MMUCC) \cite{mmucc2024} and the FHWA's Highway Safety Manual (HSM) \cite{aashto_hsm_2010}, further augmented with synthetic data. Their work demonstrated that an efficient fine-tuning approach, focusing only on the final layers, yielded a model significantly more adept at addressing safety-related queries than the base LLM. Similarly, Wang et al. \cite{wang2024transgpt} created the {TransGPT} family of models. {TransGPT-SM}, based on {ChatGLM2-6B}, was fine-tuned on an extensive corpus of transportation documents, books, and reports, showing improved performance in tasks like analyzing traffic documents and answering driving exam questions. These efforts highlight the value of injecting explicit domain knowledge through targeted fine-tuning.

Recognizing that transportation inherently involves multi-modal data, researchers are also developing {multi-modal foundation models}. Wang et al. \cite{wang2024transgpt} extended their work with {TransGPT-MM}, fine-tuning {VisualGLM-6B} on paired images (e.g., traffic signs, landmarks) and corresponding text descriptions to enhance performance on image-based traffic queries. Addressing the complex integration challenge in autonomous driving, Wang et al. \cite{wang2024bevgpt} proposed {BEVGPT}, a generative pre-trained large model framework specifically designed to unify prediction, decision-making, and motion planning. Uniquely, {BEVGPT} relies solely on bird's-eye-view (BEV) imagery as input. Pre-trained on vast real-world driving data and subsequently fine-tuned in simulation, {BEVGPT} demonstrated superior performance in decision-making accuracy and safety metrics compared to baseline methods, showcasing the potential of large transformer models adapted for visual scene understanding in vehicle control. This trend towards unified, multi-modal systems is further corroborated by Luo et al. \cite{luo2024delving} in their comprehensive survey on {Multi-modal Multi-task Visual Understanding Foundation Models (MM-VUFMs)} for road scenes. They document the shift from traditional single-task models to unified architectures leveraging {LLMs} and {VLMs} to process diverse inputs (camera, LiDAR, text, actions) and concurrently handle multiple driving tasks, thereby enabling advanced capabilities like open-world understanding, continual learning, and generative world modeling.

Cui et al. \cite{cui2025trafficllm} tackled the challenge of adapting LLMs to the unique structure of time series traffic data with {TrafficLLM}, a framework designed to overcome the modality gap between natural language and structured traffic data, improve generalization across diverse traffic analysis tasks, and reduce adaptation costs. {TrafficLLM} employs domain-specific tokenization for heterogeneous inputs, a dual-stage tuning pipeline to disentangle instruction learning from data pattern learning, and PEFT for efficient scalability. Their results indicate strong performance and robustness on traffic detection and generation tasks. Lee et al. \cite{lee2024step} focused on spatial-temporal traffic prediction with {STEP-LLM}. This framework integrates spatial-temporal-enriched prompting, using a novel {Condensed Spatial Prompting (CSP)} method to compress graph interpretations into concise text prompts, alongside patch reprogramming of traffic flow series, all processed by a frozen pre-trained {LLM}. {STEP-LLM} demonstrated superior accuracy over specialized traffic models while requiring significantly fewer computational resources, highlighting the potential of LLMs combined with clever input engineering for complex forecasting tasks.

Beyond models tailored for specific transportation applications, broader {foundational models for relevant data types} are also emerging. Rasul et al. \cite{rasul2023lag}, with {Lag-Llama}, developed a general foundation model for time series forecasting, pre-trained on a massive corpus that included traffic data. Its capability for zero-shot forecasting on unseen traffic datasets suggests that large-scale pre-training can create versatile models applicable to transportation without task-specific training. Furthermore, LLMs are being utilized as tools to {build structured domain knowledge}. Tang et al. \cite{tang2023domain} explored domain knowledge distillation, using structured prompts with {ChatGPT} to automatically construct an ontology for the autonomous driving domain, extracting concepts, hierarchies, and relationships. While requiring human oversight, this shows LLMs' potential in building structured knowledge bases, sometimes formalized using a distillation loss like minimizing the KL divergence between teacher and student output distributions $L_\text{KD} = D_{\text{KL}}(p_{\text{student}} \parallel p_{\text{teacher}})$.

In summary, the adaptation of LLMs and the development of foundation models for transportation represent a rapidly evolving frontier. Key strategies include fine-tuning on domain-specific text \cite{zheng2023trafficsafetygpt, wang2024transgpt}, integrating multi-modal data streams \cite{wang2024transgpt, wang2024bevgpt, luo2024delving}, designing specialized architectures and prompting techniques for unique data structures like traffic flow \cite{cui2025trafficllm, lee2024step}, building general foundation models for underlying data types like time series \cite{rasul2023lag}, and leveraging LLMs for knowledge base construction \cite{tang2023domain}. These efforts underscore the necessity of specialized adaptation, domain-specific representations, efficient tuning methods like {PEFT} \cite{cui2025trafficllm}, and multi-modal capabilities to fully harness the potential of large models for improving roadway safety and mobility. Open challenges remain, particularly in ensuring robustness, interpretability, and efficient operation in real-world, safety-critical systems \cite{luo2024delving}.

\subsection{Explainability, Trust, and Safety Case Generation}

A significant challenge in deploying advanced AI within safety-critical transportation systems is the inherent lack of transparency often associated with traditional deep learning models, commonly referred to as 'black boxes'. While these models can achieve high predictive accuracy, their inability to provide clear insights into their decision-making processes can impede user trust, hinder regulatory acceptance, and complicate debugging \cite{guo2024towards}. LLMs, with their sophisticated natural language processing and generation capabilities, offer a promising avenue to address this challenge, potentially paving the way for more explainable, trustworthy, and ultimately safer AI applications in roadway safety and mobility enhancement.

Several recent studies have explored leveraging LLMs to enhance interpretability in various transportation contexts. Guo et al. \cite{guo2024towards} introduced {xTP-LLM}, a framework specifically designed for \textit{explainable} traffic flow prediction. Their approach uniquely transforms multi-modal inputs—spanning historical traffic data, spatial context from Points of Interest (PoIs), weather conditions, and temporal factors—into structured textual prompts. By fine-tuning an LLM (Llama2) and employing careful prompt engineering that incorporates domain knowledge and CoT reasoning, {xTP-LLM} not only generates accurate traffic forecasts competitive with state-of-the-art deep learning methods but also outputs coherent, natural language explanations justifying the predictions based on the input factors. This dual output significantly enhances interpretability and demonstrated robustness across diverse conditions \cite{guo2024towards}.

Extending the concept of explainability to driver behavior analysis and autonomous driving decisions, researchers have utilized LLMs to articulate the reasoning behind system actions. Zhang et al. \cite{zhang2024integrating}, developed {DDLM} for distracted driving classification, incorporating a reasoning chain framework to make the LLM's classification and risk assessment process more transparent through step-by-step explanations. Similarly, the {GPT-Driver} model proposed by Mao et al. \cite{mao2023gpt} generates not only driving waypoints but also accompanying natural language CoT explanations for maneuvers, such as justifying deceleration due to a leading vehicle. Yildirim et al. \cite{yildirim2024highwayllm} adopted a comparable approach with their {HighwayLLM}, which provides natural language justifications for its predicted trajectories. Addressing explainability through direct interaction, Marcu et al. \cite{marcu2024lingoqa} developed the {LingoQA} dataset and framework, enabling Visual Question Answering (VQA) focused on understanding Autonomous Vehicle (AV) behavior, specifically allowing users to pose "why" questions. Furthermore, Jain et al. \cite{jain2024semantic} explored Large Vision-Language Models (LVLMs) like GPT-4 and Video-LLaVA for comprehensive semantic understanding of complex traffic scenes by integrating visual and LiDAR data, demonstrating their ability to provide contextually relevant answers to critical transportation-related queries \cite{jain2024semantic}. Hussien et al. \cite{hussien2025rag} employed RAG combined with knowledge graphs, allowing LLMs to generate grounded natural language explanations for pedestrian and driver behavior predictions derived from the structured knowledge base.

Beyond explaining model predictions or actions, LLMs are also being investigated for their potential in formal safety assurance processes. Sivakumar et al. \cite{sivakumar2024prompting} explored the challenging task of automatically generating {safety cases}—structured arguments required to demonstrate that a system meets acceptable safety levels—using {GPT-4 \cite{liu2023visual}} and Goal Structuring Notation (GSN). Their findings indicated that while {GPT-4 \cite{liu2023visual}} demonstrated a good grasp of GSN principles and could generate moderately accurate safety cases, particularly when guided by domain knowledge and syntax prompts (achieving high semantic similarity scores of 0.86-0.89), the structural correctness remained around 80-90\%. This suggests that LLMs can serve as valuable assistants in the complex and rigorous process of safety case generation but cannot yet fully replace human expertise and oversight, especially given the critical importance of formal safety assurance in transportation \cite{sivakumar2024prompting}.

\subsection{Edge Computing and Lightweight Models}

The demanding requirements of real-time roadway safety and mobility applications, necessitate low latency and computational efficiency. Deploying LLMs in these contexts often involves resource-constrained edge devices like vehicles or Roadside Units (RSUs). Consequently, significant research focuses on two complementary strategies: leveraging edge computing infrastructure and developing or adapting lightweight LLM architectures to make them suitable for deployment outside of powerful cloud servers.

Edge computing brings computation closer to the data source, reducing communication delays inherent in cloud-based processing. Huang et al. \cite{huang2024efficient} explored this by deploying LLMs on interconnected RSUs using 5G for communication. Their framework enables localized data processing and rapid hazard dissemination. By employing a multi-modal prompt strategy incorporating environmental, agent, and motion data, they significantly enhanced the accuracy of LLM-generated driving scene narration (78.2\% with Video-ChatGPT) and reasoning (81.7\%), while achieving substantially faster response times compared to traditional methods, demonstrating the suitability of edge-based LLMs for real-time analysis. Similarly, the work by Rong et al. \cite{rong2024edge} on STGLLM-E (Spatio-Temporal Generative Large Language Model on Edge) specifically targets large-scale traffic flow prediction within 6G-IATS. Their approach utilizes a tailored edge-computing strategy, offloading tasks from a central server to distributed edge clusters. This, combined with graph segmentation techniques preserving road connectivity and attention mechanisms capturing spatio-temporal dependencies, aims to reduce computational bottlenecks and improve real-time performance. The survey by Panyam et al. \cite{panyam2025survey} further underscores the importance of hybrid cloud-edge architectures in V2X systems to effectively balance latency constraints with computational demands.

Complementary to edge deployment, optimizing the LLMs themselves for efficiency is crucial. This involves creating inherently lightweight models or adapting larger models using various techniques. Rong et al. introduced LSGLLM-E (Lightweight Spatio-temporal Generative Large Language Model on Edges) \cite{rong2024large}, which integrates a specialized spatio-temporal module (STM) with a pre-trained LLM backbone (GPT-2 \cite{radford2019language}) and leverages edge intelligence for distributed computation, improving both accuracy and efficiency in large-scale prediction tasks. Another approach involves using LLMs for tasks that enhance system efficiency, such as data compression. Yang et al. \cite{yang2024transcompressor} proposed TransCompressor, where sensor data is compressed locally (e.g., via skip sampling) and then reconstructed by a cloud-based LLM (like GPT-4) using zero-shot prompting, significantly improving data storage and transmission efficiency without task-specific fine-tuning.

Several parameter-efficient fine-tuning (PEFT) techniques are widely adopted to reduce the computational burden of adapting large models. Low-Rank Adaptation (LoRA) has been frequently employed \cite{lan2024traj, guo2024towards, ren2024tpllm, zhang2024integrating, pu2024autorepo, fan2024learning, onsu2025leveraging}. LoRA is based on the hypothesis that the change in weights ($\Delta W$) needed for task adaptation has a low "intrinsic rank." It freezes the original weights ($W$) and injects trainable rank decomposition matrices ($B$ and $A$) into specific layers (e.g., attention layers). The update is approximated as $\Delta W = BA$, where $W \in \mathbb{R}^{d \times k}$, $B \in \mathbb{R}^{d \times r}$, $A \in \mathbb{R}^{r \times k}$, and the rank $r \ll \min(d, k)$. This reduces the number of trainable parameters for the update from $d \times k$ to $r(d+k)$, which is substantially smaller for small $r$.

{Quantization} is another critical technique for creating lightweight models. It involves reducing the numerical precision of the model's weights and/or activations (e.g., from 32-bit floating-point, FP32, to 8-bit integer, INT8, or even lower). Conceptually, a high-precision weight tensor $W$ is approximated by a low-precision tensor $W_q$ using a scaling factor $S$ (and potentially a zero-point $Z$), such that $W \approx S \cdot (W_q - Z)$. This reduces the model's memory footprint and can accelerate computation on hardware supporting low-precision arithmetic. Techniques like QLoRA \cite{pu2024autorepo, abdelrahman2024video} combine quantization (e.g., 4-bit NormalFloat, NF4) with LoRA, further optimizing memory usage during fine-tuning.

Other strategies include keeping the core LLM frozen and only training lightweight input/output layers, as demonstrated in the TIME-LLM framework \cite{jin2023time}. Furthermore, selecting smaller yet capable base models like LLaMA-7B \cite{zheng2023trafficsafetygpt}, Phi-3 mini \cite{abdelrahman2024video}, or specialized efficient models like LightGPT \cite{lai2023large} inherently reduces resource requirements.

Despite these advancements, challenges related to inference latency persist, particularly for safety-critical, real-time control loops. HighwayLLM \cite{yildirim2024highwayllm} noted significant latency (2.9 - 6.8 seconds) for LLM-based planning and safety checks compared to traditional Reinforcement Learning approaches (2ms). Similarly, LA-Light \cite{wang2024llm} highlighted latency concerns arising from frequent interactions with the LLM. Addressing end-to-end LLM serving efficiency is also vital. Frameworks like ScaleLLM \cite{yao2024scalellm}, though not specific to transportation, focus on optimizing the entire serving pipeline through techniques like efficient routing, model parallelization, quantization, batching, and optimized inference engines. Such optimizations are crucial for deploying LLMs at scale, whether on the edge or in supporting cloud infrastructure for transportation services.

\section{Challenges and Future Directions}

Despite the promising advancements surveyed in this review, the effective and responsible deployment of LLMs in roadway safety and mobility enhancement faces significant challenges. Addressing these limitations and exploring targeted future research avenues is crucial for realizing the full potential of this technology in the complex and safety-critical transportation domain.

\subsection{Addressing Inherent LLM Limitations}
While powerful, current LLMs possess inherent limitations that are particularly concerning for transportation applications.
\subsubsection{Hallucination and Factual Inconsistency:} LLMs are generative models, fundamentally designed to predict probable sequences of text based on patterns learned from vast datasets \cite{kenthapadi2024grounding, huang2025survey}. A significant limitation arising from this is "hallucination," where the model generates text that is plausible-sounding but factually incorrect, unfaithful to provided source documents, or nonsensical \cite{kenthapadi2024grounding}. In transportation, this could manifest critically as citing non-existent traffic regulations, predicting physically impossible traffic dynamics, generating unsafe driving maneuvers, or fabricating details in accident report summaries \cite{mumtarin2023large}. Conceptually, these models optimize for likelihood $P(\text{output} | \text{input})$ within their learned distribution, which does not inherently guarantee factual correctness. While techniques like RAG aim to mitigate this by conditioning generation on externally retrieved knowledge \cite{hussien2025rag, luo2024deciphering, you2025comprehensive}, challenges persist in ensuring the relevance and accuracy of retrieved information and the model's faithfulness in utilizing it \cite{kenthapadi2024grounding, huang2025survey}. Industrial benchmarks focusing on traffic incident data confirm significant hallucination issues, especially regarding temporal accuracy \cite{liindustrial}. Establishing robust mechanisms for fact-checking and ensuring verifiable accuracy is paramount before deploying LLMs in safety-critical transportation functions.

\subsubsection{Grounding in Dynamic Physical Reality:} Transportation systems are intrinsically linked to the physical world, operating within constantly changing environments. A core challenge, often related to the "symbol grounding problem" in AI, is effectively connecting the symbolic representations learned by LLMs from text to the continuous, noisy, and often incomplete data streams from real-world sensors (e.g., cameras, LiDAR, GPS, loop detectors) \cite{kenthapadi2024grounding}. LLMs must be able to interpret and reason over this dynamic sensory input, discerning meaningful patterns from noise, handling conflicting data sources, and updating their understanding of the world state \cite{zhang2024grounding}. Over-reliance on the model's parametric knowledge (learned during training) can lead to outputs inconsistent with current reality, while overly trusting potentially faulty sensor data is equally problematic. Developing robust techniques for grounding LLM reasoning in uncertain, high-dimensional, and dynamic spatio-temporal data streams is an open research problem \cite{zhang2024grounding}. This involves integrating probabilistic reasoning or uncertainty quantification methods to manage imperfect information.

\subsubsection{Numerical Reasoning and Optimization Inadequacies:} Many critical transportation tasks require precise quantitative reasoning, adherence to physical laws expressed mathematically, or solving optimization problems. Examples include calculating optimal traffic signal timings (often formulated as minimizing delay subject to constraints, e.g., $\min \sum \text{delay}_i \text{ s.t. phase constraints}$) \cite{dai2024large}, generating accurate vehicle trajectories defined by coordinates \cite{lan2024traj}, or performing complex risk assessments involving probabilistic calculations. LLMs, primarily trained on text, often exhibit weaknesses in performing accurate multi-step arithmetic, symbolic manipulation, or executing optimization algorithms \cite{syed2024benchmarking}. Hybrid systems, where the LLM acts as a high-level planner or translates natural language requests into formal problem specifications, while delegating the precise calculations or optimization to dedicated solvers (e.g., mathematical programming solvers, physics simulators, specialized controllers) \cite{sha2023languagempc, zhang2024trafficgpt, da2024open}, represent a practical workaround. However, the seamless and reliable integration of these components remains a challenge.

\subsubsection{Scalability, Efficiency, and Latency Constraints:} State-of-the-art LLMs contain billions of parameters, requiring substantial computational resources (measured in FLOPs - floating-point operations) and energy for both training and inference \cite{mao2023gpt, wang2024llm, panyam2025survey, jin2023time}. The inference latency (time taken to generate a response) can be significant, potentially exceeding the strict real-time requirements (often milliseconds) of applications like autonomous vehicle control, collision avoidance systems, or adaptive traffic signal adjustments \cite{yildirim2024highwayllm, lai2023large}. The computational complexity of the attention mechanism in Transformers, often scaling quadratically ($O(n^2)$) with input sequence length $n$, exacerbates this for long contexts. Addressing the trade-off between model capability (often correlated with size) and operational efficiency (latency, throughput, cost) is crucial \cite{yildirim2024highwayllm, lai2023large}. Further research focuses on techniques like PEFT methods \cite{lan2024traj, guo2024towards}, model quantization \cite{pu2024autorepo, abdelrahman2024video}, pruning (removing redundant parameters), knowledge distillation \cite{tang2023domain}, deploying models on edge devices closer to sensors \cite{rong2024large, rong2024edge}, and developing optimized inference engines and serving frameworks \cite{yao2024scalellm}. These are essential for making LLMs viable in resource-constrained and time-sensitive transportation environments.

\subsubsection{Consistency and Reliability Concerns:} The inherent stochasticity in LLM generation, often controlled by decoding strategies like temperature sampling (where output probability $P(token_i) \propto \exp(\frac{logit_i}{T})$ is influenced by temperature $T$), can lead to non-deterministic outputs; the same input may yield different responses across multiple queries \cite{mumtarin2023large, tang2023domain}. While beneficial for creative tasks, this lack of consistency poses a significant risk for safety-critical transportation systems that demand predictable and reliable behavior \cite{lai2023large}. Ensuring that an LLM-based system behaves consistently and dependably under identical conditions is vital. Strategies to improve reliability include constrained decoding (forcing outputs to adhere to specific rules or formats), employing self-consistency mechanisms (generating multiple outputs and selecting the most consistent one) \cite{cai2025text2scenario}, and implementing rigorous verification and validation (V\&V) protocols, although applying traditional V\&V to complex, non-deterministic AI remains challenging \cite{syed2024benchmarking}.

\subsection{Data Requirements, Privacy, and Bias Concerns}
The efficacy, reliability, and societal acceptance of LLMs in the transportation domain are fundamentally intertwined with the characteristics of the data used for their training, fine-tuning, and grounding. Addressing challenges related to data availability, inherent biases, and privacy protection is paramount for responsible innovation

\subsubsection{Data Availability, Quality, Diversity, and Cost:} LLMs are data-hungry models whose performance scales with the volume and quality of training data. Their ability to learn complex patterns, understand nuanced transportation scenarios, and crucially, generalize to unseen situations depends heavily on exposure to vast, diverse datasets \cite{su2024large, kuftinova2024large, jaradat2025leveraging}. Effective transportation LLMs require data encompassing the full spectrum of operational conditions, including routine traffic flow, congestion patterns, diverse weather impacts, infrastructure variations, and, critically, rare but high-impact events such as crashes, near-misses, and emergency vehicle interactions. High-quality data in this context implies not only accuracy, completeness, and semantic richness but also temporal relevance, geospatial coverage, and data integrity (ensuring data hasn't been tampered with, guarding against potential data poisoning attacks that could compromise LLM safety). Creating and maintaining such domain-specific corpora (e.g., CrashEvent \cite{fan2024learning}, TrafficSafety-2K \cite{zheng2023trafficsafetygpt}, Walk-the-Talk \cite{ramesh2024walk}, LingoQA \cite{marcu2024lingoqa}, TransEval \cite{wang2024transgpt}, TransportBench \cite{syed2024benchmarking}) is a resource-intensive undertaking, involving significant costs not only for data acquisition (e.g., sensor deployment, data storage) but also for data processing and annotation (e.g., labeling events, geocoding text descriptions), which are often required for supervised fine-tuning. Data scarcity, particularly for specific geographic regions, underrepresented demographic groups, or specific low-frequency event types (the "long tail" of events), significantly hinders model robustness, limits the effectiveness of few-shot or zero-shot learning paradigms, and can lead to models that perform poorly outside their training distribution \cite{ren2024tpllm, das2023classifying, jin2023time}.

\subsubsection{Temporal Dynamics and Concept Drift:} Transportation systems are inherently dynamic; traffic patterns evolve daily and seasonally, infrastructure changes, travel behaviors shift (e.g., adoption of new modes, impact of policies), and external events (e.g., pandemics, major construction) alter system dynamics. Data used to train LLMs represents a snapshot in time. As the real-world system evolves, the statistical properties of the incoming data may change, leading to  concept drift . An LLM trained on historical data may gradually lose its predictive accuracy or relevance if not continuously monitored and updated. This necessitates strategies for ongoing data collection, model performance monitoring, and  continuous learning or periodic retraining  to ensure the LLM adapts to the evolving transportation landscape and maintains its reliability over time.

\subsubsection{Algorithmic Bias and Fairness:} A significant ethical challenge lies in the potential for LLMs to inherit and amplify societal biases present in transportation data. This phenomenon, often termed algorithmic bias, can arise from various sources:

\begin{itemize}
    \item Sampling Bias: Data collection mechanisms might over-represent certain areas or populations while under-representing others.
    \item Measurement Bias: Proxies used for safety or mobility (e.g., reported crashes vs. actual risk) may themselves be biased.
    \item Societal Bias: Historical inequities reflected in infrastructure allocation, policing patterns, or differing mobility needs across demographics can be encoded in the data \cite{kuftinova2024large, yang2024applying, qasemi2023traffic}.
\end{itemize}
LLMs trained on such data risk perpetuating these biases, potentially leading to inequitable outcomes. Examples include unfair allocation of green time at signalized intersections disadvantaging certain user groups, safety interventions that disproportionately target specific communities, or mobility-as-a-service recommendations that neglect underserved areas. Addressing this requires developing techniques for bias detection (e.g., using statistical parity, equal opportunity, or counterfactual fairness metrics adapted for transportation contexts) and bias mitigation (e.g., data augmentation, re-weighting, adversarial debiasing, or incorporating fairness constraints during model training). Defining and achieving "fairness" in the complex socio-technical system of transportation remains an open research question.

\subsubsection{Privacy Protection in Sensitive Data Environments:} Transportation data, by its nature, is often highly sensitive, containing Personally Identifiable Information (PII) or data from which sensitive attributes can be inferred. This includes precise vehicle trajectories (revealing home locations, workplaces, and routines), video feeds from infrastructure cameras or vehicles (capturing faces, license plates, pedestrian activity), and user interaction logs or feedback \cite{kenthapadi2024grounding, panyam2025survey}. Using LLMs to process this data raises significant privacy concerns, as models could potentially memorize sensitive training data excerpts or enable inference attacks that reveal private information. This is further complicated by data governance issues, concerning who owns, controls, and grants access to data collected from diverse public and private sources. Robust Privacy-Preserving Techniques (PPTs) are therefore essential. Key approaches include:
\begin{itemize}
    \item Data Minimization and Anonymization: Collecting only necessary data and applying techniques to remove or obfuscate direct identifiers, although true anonymization of high-dimensional data like trajectories is notoriously difficult.
    \item Novel Data Representations: Transforming raw data into less sensitive formats, such as the abstract textual descriptions of traffic events used in VTPM \cite{abdelrahman2024video}, reducing direct exposure of raw visual or trajectory data.
    \item Differential Privacy (DP): A formal framework providing mathematical guarantees on privacy. By adding calibrated statistical noise to data, computations, or model outputs, DP ensures that the presence or absence of any single individual's data in the dataset has a provably small effect on the outcome. This is often characterized by a privacy budget parameter, $\epsilon, \text{where smaller } \epsilon \text{ implies stronger privacy.}$ 
    \item Federated Learning (FL): A decentralized machine learning approach where the model is trained across multiple local datasets (e.g., on vehicles or edge devices) using local data without exchanging raw data. Only aggregated model parameters or updates are shared with a central server, enhancing data locality and privacy \cite{tian2023vistagpt}.
\end{itemize}
    
Implementing these techniques often involves a fundamental privacy-utility trade-off: stronger privacy guarantees may reduce data utility and negatively impact model accuracy or the granularity of insights. Finding the optimal balance for specific transportation applications, considering both ethical mandates and performance requirements within existing data governance frameworks, is a critical ongoing challenge.

    \subsection{Real-world Deployment and Integration Challenges}
Transitioning LLM-driven applications from the controlled confines of research environments to the dynamic and safety-critical complexities of real-world transportation systems presents substantial hurdles. Addressing these challenges is paramount for realizing the potential benefits of LLMs in enhancing roadway safety and mobility.

    \subsubsection{The Sim-to-Real Gap: Bridging Simulated Training and Real-World Dynamics}
A significant challenge lies in the \textit{sim-to-real gap}, which refers to the discrepancy in performance observed when models trained or validated primarily in simulation are deployed in the real world. This gap arises because simulations, while valuable, often simplify or imperfectly model complex real-world phenomena. Key contributing factors include mismatches in:
    \begin{itemize}
        \item \textbf{System Dynamics:} Differences between simulated and actual vehicle physics, powertrain responses, sensor characteristics (noise profiles, biases, failure modes), and the stochastic nature of multi-agent interactions (e.g., human driver reactions, pedestrian behaviors) \cite{da2023llm}.
        \item \textbf{Environmental Variability:} Inability of simulations to capture the full spectrum of real-world environmental conditions (diverse weather, lighting changes, road surface variations, sensor occlusions).
    \end{itemize}
Models optimized for simulated environments may fail to generalize, leading to unreliable or unsafe behavior upon deployment. Fundamentally, this is a problem of \textit{domain shift}, where the data distribution of the target domain (real world, $\mathcal{P}_T$) differs from the source domain (simulation, $\mathcal{P}_S$). Closing this gap necessitates robust \textit{domain adaptation} and \textit{transfer learning} techniques. The objective is to learn a model $f$ that performs well on the target domain, often by minimizing some measure of divergence between the domains or learning domain-invariant features. Strategies include:
    \begin{itemize}
        \item \textbf{Advanced Transfer Learning Methods:} Techniques like PromptGAT, which adapts graph attention networks using prompts, aim to transfer knowledge effectively from simulation \cite{da2023llm}.
        \item \textbf{Domain Randomization:} Intentionally varying simulation parameters (e.g., friction coefficients, sensor noise levels, lighting conditions) during training. This exposes the model to a wider range of conditions, aiming to create a simulation distribution that implicitly covers the target real-world distribution, thus improving robustness. Mathematically, parameters $\theta$ are sampled from a distribution $P(\theta)$ to generate diverse training scenarios.
        \item \textbf{Rigorous Real-World Validation:} Extensive testing and fine-tuning using real-world data are indispensable to validate performance and adapt models post-simulation \cite{lai2023large}. This often involves costly data collection and iterative refinement cycles.
    \end{itemize}

    \subsubsection{Interoperability and Seamless Integration within Heterogeneous Ecosystems}
Real-world transportation systems constitute a complex \textit{System of Systems (SoS)}, characterized by heterogeneity in both hardware and software components. Integrating LLM-based functionalities seamlessly into this existing infrastructure poses significant \textit{interoperability} challenges. Interoperability, in this context, is the ability of different systems and components—often from various manufacturers and developed over different timespans—to exchange information and use the information that has been exchanged effectively. Key aspects include:
    \begin{itemize}
        \item \textbf{Hardware Diversity:} LLMs need to interact with a wide array of physical devices, including vehicle sensors (cameras, LiDAR, radar), actuators (braking, steering systems), roadside units (RSUs), traffic controllers, and diverse vehicle Electronic Control Units (ECUs) with varying computational capabilities.
        \item \textbf{Software and Communication Complexity:} Integration involves interfacing with legacy software systems, diverse databases, proprietary communication protocols, and standardized protocols (e.g., V2X communication standards like DSRC or C-V2X based on SAE J2735 or ETSI standards, in-vehicle networks like CAN bus, automotive middleware like AUTOSAR).
    \end{itemize}
Achieving seamless integration requires adherence to \textit{Systems Engineering} principles, focusing on well-defined interfaces and modularity. The adoption of \textit{standardized data formats and communication protocols} is crucial \cite{panyam2025survey, da2024open}. Furthermore, architectural patterns like \textit{Service-Oriented Architecture (SOA)} or \textit{microservices} can provide the flexibility needed to incorporate new LLM-based services without disrupting the entire system. Middleware solutions can also play a role in abstracting hardware and communication heterogeneity.

    \subsubsection{Ensuring Robustness and Resilience in Safety-Critical Operations}
Transportation applications, particularly those related to safety, demand high levels of \textit{robustness} and \textit{resilience}. Robustness refers to the ability of the system to maintain performance under uncertainty, disturbances, or perturbations, while resilience refers to the system's ability to detect, respond to, and recover from failures or adverse events. Real-world systems must contend with:
    \begin{itemize}
        \item \textbf{Component Failures:} Sensor malfunctions (noise, bias, complete failure) \cite{wang2024llm}, actuator faults, computational errors.
        \item \textbf{Communication Disruptions:} Intermittent connectivity, packet loss, or high latency, especially critical in Vehicle-to-Everything (V2X) communication scenarios \cite{you2024v2x}.
        \item \textbf{Environmental Disturbances:} Unpredictable conditions like severe weather, sudden changes in lighting, or road debris not encountered during training.
        \item \textbf{Adversarial Attacks:} Intentional efforts to compromise system behavior by manipulating inputs. For LLMs, this includes \textit{prompt injection} attacks or crafting adversarial sensor inputs designed to mislead the model's perception or decision-making \cite{kenthapadi2024grounding}. An attacker might seek a minimal perturbation $\delta$ added to an input $x$ (e.g., sensor data or text prompt) such that the LLM output $f(x+\delta)$ leads to an unsafe or undesirable outcome.
    \end{itemize}
Ensuring dependable operation necessitates incorporating principles of \textit{fault tolerance}. This involves designing systems with \textit{graceful degradation} capabilities (maintaining essential functions, possibly at reduced performance, upon partial failure) and robust \textit{fail-safe} or \textit{fail-operational} mechanisms to ensure safety even when critical components malfunction. Techniques from \textit{robust control theory} may also be relevant for designing controllers that remain stable despite uncertainties and disturbances.

\subsection{Ethical Considerations and Trust}
The deployment of LLMs in safety-critical transportation applications raises fundamental ethical questions.

\subsubsection{Safety Assurance and Validation:} How can we rigorously verify and validate the safety and reliability of LLM-driven systems, especially given their non-deterministic nature and potential for hallucination? Traditional software verification methods may be insufficient. Formal methods, extensive scenario-based testing \cite{cai2025text2scenario}, and potentially the generation of formal safety cases \cite{sivakumar2024prompting} are necessary. Human oversight remains crucial, particularly in the initial stages of deployment \cite{manas2024tr2mtl, tang2023domain, wang2024leveraging}.
\subsubsection{Bias and Fairness:} Ensuring that LLM-based systems do not perpetuate or create societal inequities is vital. This requires careful auditing of data and algorithms, fairness-aware design, and consideration of impacts on different user groups and communities. For instance, traffic signal optimization should not consistently disadvantage pedestrians or certain neighborhoods.
\subsubsection{Accountability and Liability:} Establishing clear lines of responsibility when an LLM-driven system contributes to an accident or causes harm is a complex legal and ethical challenge that needs resolution before widespread deployment.
\subsubsection{Transparency and Explainability:} While LLMs can generate explanations \cite{guo2024towards, zhang2024integrating, mao2023gpt, hussien2025rag}, ensuring these explanations are faithful to the model's actual reasoning process and truly informative for users (drivers, operators, investigators) is an ongoing research area \cite{kenthapadi2024grounding}. Building trust requires transparent and understandable systems \cite{lai2023large, marcu2024lingoqa}.

\subsection{Promising Future Research Avenues}
Addressing the multifaceted challenges outlined in the preceding sections necessitates concerted and innovative research efforts across multiple domains. Building upon the reviewed literature, the following avenues represent particularly promising directions for advancing the application of LLMs in the enhancement of roadway safety and mobility.

    \subsubsection{Advanced Multimodal Integration and Fusion}
    Future work should focus on progressing beyond relatively simple concatenation or early fusion of vision and language data. This involves the development of sophisticated fusion architectures capable of integrating information from a wider array of heterogeneous sensors, including radar, LiDAR, thermal, ultrasonic, IMU, audio, proprioceptive sensors, and V2X messages. Key challenges include handling asynchronous data streams, effectively modeling cross-modal correlations, robustly managing sensor noise and uncertainty, and achieving true semantic understanding across modalities \cite{jain2024semantic, wang2024transgpt, qasemi2023traffic, su2024large}. Research into attention mechanisms that operate across diverse data types and techniques for learning joint embeddings that capture rich cross-modal interactions will be vital for achieving comprehensive situational awareness \cite{kenthapadi2024grounding, you2025comprehensive, wang2023accidentgpt, you2024v2x}.

    \subsubsection{Native Spatio-Temporal Reasoning Capabilities}
    Current LLMs primarily excel at processing discrete sequences of tokens. Enhancing their ability to inherently understand and reason about continuous space and time is crucial for transportation tasks involving trajectories, dynamic interactions, and physical constraints. Future architectural designs might integrate principles from Geometric Deep Learning, Graph Neural Networks (GNNs) for modeling road networks and agent interactions \cite{rong2024edge}, Physics-Informed Machine Learning (PIML) to incorporate domain knowledge about dynamics, or continuous state-space models \cite{rong2024large, li2024urbangpt, liu2024spatial, lan2024traj}. The objective is to develop models that can naturally represent and predict spatio-temporal phenomena without relying solely on discretized representations.

    \subsubsection{Causal Inference and Counterfactual Reasoning}
    To move beyond correlation-based predictions towards a deeper understanding and reliable decision-making, LLMs require enhanced capabilities for causal inference. This involves not only predicting what is likely to occur but also understanding why it occurs and what would transpire under different circumstances (counterfactuals). Research should explore the integration of LLMs with frameworks such as Structural Causal Models (SCMs) or the potential outcomes framework, enabling them to model interventions (e.g., estimating the safety impact of a new traffic policy) and answer complex counterfactual questions ("What would have happened if the vehicle had braked 0.5 seconds earlier?") \cite{fan2024learning}. This is essential for robust safety analysis, planning, and policy evaluation.

    \subsubsection{Synergistic Human-AI Collaboration and Interaction}
    Designing effective paradigms for collaboration between LLMs and human users—such as drivers, traffic operators, fleet managers, and emergency responders—is critical. This extends beyond simple query-response interfaces (e.g., ChatSUMO \cite{li2024chatsumo} or Mobility ChatBot \cite{padoan2024mobility}) to encompass principles from Human-Computer Interaction (HCI), Cognitive Systems Engineering, and Shared Autonomy. Research should focus on developing intuitive and adaptive interfaces, mechanisms for effective trust calibration (ensuring appropriate reliance on the AI), methods for shared control and decision-making, and clear communication protocols that leverage the complementary strengths of humans (intuition, ethics, complex reasoning) and LLMs (data processing, pattern recognition) \cite{dai2024large, wang2023accidentgpt, wang2024leveraging}.

    \subsubsection{Continuous Learning, Adaptation, Online Refinement, and Memory}
    Transportation environments are inherently non-stationary; traffic patterns evolve, infrastructure changes, and sensor characteristics may drift. LLMs deployed in this domain must be capable of continuous or lifelong learning, adapting to these changes from real-world data streams without catastrophic forgetting of previously learned knowledge. Research is needed in online learning algorithms suitable for large models, techniques to mitigate catastrophic forgetting, and efficient methods for model updating. Furthermore, integrating effective memory mechanisms \cite{fang2025towards}, allowing models to explicitly store, retrieve, and reason about past experiences (e.g., near-miss incidents, recurring congestion patterns), is crucial for improving performance over time.

    \subsubsection{Verified Explainability, Trustworthiness, and Safety Verification}
    Building trust requires not only generating explanations \cite{guo2024towards, hussien2025rag} but also ensuring their \textit{faithfulness} (accuracy in reflecting the model's reasoning) and developing methods for the formal verification of safety properties. Research should focus on enhancing XAI techniques for transportation contexts and exploring the application of \textit{formal methods} or \textit{runtime verification} techniques to provide provable guarantees (even if limited to specific properties or subsystems) about the behavior of LLM-driven components \cite{sivakumar2024prompting}.

    \subsubsection{Efficient Domain-Specific Foundation Models}
    While large, general-purpose LLMs provide a strong base, developing \textit{foundation models} specifically pre-trained or extensively fine-tuned on high-quality, diverse transportation datasets (text, trajectories, sensor readings, simulation logs) holds significant promise \cite{zheng2023trafficsafetygpt, wang2024transgpt, rasul2023lag}. Such domain specialization could lead to models that are not only more capable on specific transportation tasks but also potentially smaller, faster, and more data-efficient. Research into effective pre-training objectives and techniques for transportation LLMs is warranted.

    \subsubsection{Hybrid AI Architectures: LLMs + Symbolic Reasoning/Optimization}
    Exploring \textit{neuro-symbolic AI} approaches that synergistically combine the strengths of LLMs (handling unstructured data, flexible pattern recognition, natural language interaction) with traditional symbolic AI (logical reasoning, knowledge representation, planning) and mathematical optimization \cite{sha2023languagempc} is a promising frontier. For example, LLMs could generate candidate solutions or heuristics for optimization problems, interpret complex scenarios to set up constraints for symbolic reasoners, or translate optimization results into natural language explanations. This hybrid approach could lead to systems that are more robust, verifiable, and data-efficient.

    \subsubsection{Optimized and Hardware-Aware Edge Deployment}
    Realizing the potential of LLMs in vehicles (ECUs) and roadside infrastructure (RSUs) requires significant advances in deploying these models efficiently on resource-constrained edge devices. Research must focus on aggressive \textit{model compression} (quantization, pruning, distillation), automated \textit{Network Architecture Search (NAS)} targeting edge hardware profiles, efficient attention mechanisms, and \textit{hardware-software co-design} \cite{rong2024edge, yao2024scalellm}. The goal is to achieve minimal latency and power consumption while preserving task performance, adhering to the principles often associated with \textit{TinyML} \cite{warden2019tinyml}.

By strategically investing research efforts in these key areas, the community can progressively overcome current limitations and responsibly unlock the transformative potential of Large Language Models to contribute to significantly safer, more efficient, equitable, and intelligent transportation systems for the future.

\section{Conclusion}
This paper has presented a comprehensive review examining the rapidly evolving landscape of Large Language Models (LLMs) and their applications in enhancing roadway safety and mobility. Through a systematic analysis, it has been demonstrated how these powerful AI models, initially designed for natural language processing, are being adapted, customized, and deployed to address some of transportation's most pressing challenges. From improving traffic flow prediction and optimization to enhancing crash analysis and risk assessment, LLMs exhibit remarkable potential to transform multiple facets of contemporary transportation systems.

Our examination of foundational concepts has revealed how researchers are actively working to bridge the inherent modality gap between language-centric models and transportation's predominantly spatio-temporal and physical data. This is being achieved through innovative architectural adaptations, including the integration of specialized spatio-temporal modules, novel tokenization strategies, and parameter-efficient fine-tuning techniques. These advancements enable LLMs to better process and reason about the unique data characteristics inherent in transportation systems. Such adaptations, when combined with domain-specific prompting methods and multimodal integration approaches, form crucial building blocks for the effective and reliable deployment of LLMs in safety-critical transportation contexts.

In the mobility domain, significant progress has been documented across a range of applications, including traffic flow prediction and forecasting, traffic signal control optimization, trip planning, simulation, and human mobility pattern analysis. LLMs have demonstrated particular strengths in providing natural language interfaces to complex transportation systems, incorporating diverse contextual factors into predictions, enabling few-shot and zero-shot learning for data-scarce scenarios, and generating realistic simulation environments. These capabilities address long-standing challenges in transportation planning and operations, potentially leading to more efficient, adaptive, and user-centric mobility systems.

The safety domain has witnessed equally promising developments, with LLMs contributing to enhanced crash data analysis, driver behavior assessment, pedestrian safety modeling, and the formalization of traffic rules. The ability of these models to extract meaningful insights from unstructured textual narratives, integrate multimodal sensor data, provide interpretable reasoning chains, and detect near-miss incidents offers new pathways for proactive safety enhancement. By enabling a more nuanced understanding of safety-critical scenarios and supporting human decision-making processes, LLMs could contribute significantly to reducing roadway fatalities and injuries.

Enabling technologies and frameworks, such as V2X communication integration, the development of domain-specific foundation models, techniques for explainability, and optimizations for edge computing, provide the necessary infrastructure for practical deployment. These cross-cutting developments address crucial implementation challenges, including computational efficiency, reliability, and trustworthiness, which must be overcome for widespread real-world adoption.

Despite the significant promise demonstrated, substantial challenges remain that require further investigation. Inherent limitations of LLMs, including issues of hallucination, deficits in native spatio-temporal reasoning, inadequacies in numerical computation, and efficiency constraints, necessitate continued research attention. Furthermore, data requirements, privacy considerations, potential algorithmic biases, and the complexities of integration with legacy systems pose additional hurdles. Most critically, ensuring robust validation, verification, and safety assurance in these complex, often black-box systems remains an ongoing challenge that must be rigorously addressed before widespread deployment in safety-critical transportation functions.

Looking forward, several research directions appear particularly promising for future work. These include advanced multimodal integration, enhanced native spatio-temporal reasoning capabilities within LLMs, improved causal inference, the development of synergistic human-AI collaboration frameworks, mechanisms for continuous learning and adaptation, privacy-preserving distributed learning techniques, and the establishment of transportation-specific benchmarking standards. Concurrent advances in edge computing, model compression, and hardware acceleration will further enable efficient deployment in resource-constrained transportation environments.

As LLM technology continues its rapid evolution, an increasing integration with transportation systems across multiple scales—from individual vehicles to citywide networks—is anticipated. This integration holds the potential to create more intelligent, responsive, and human-centered transportation systems that enhance both safety and mobility. However, realizing this potential will require sustained interdisciplinary collaboration among transportation engineers, computer scientists, human factors researchers, policymakers, and ethicists to ensure these powerful tools are deployed responsibly, equitably, and effectively.

\ifCLASSOPTIONcaptionsoff
  \newpage
\fi



\bibliographystyle{IEEEtran}
\bibliography{ref}




\end{document}